\pdfoutput=1

\documentclass[11pt]{article}

\usepackage[final]{acl}

\usepackage{times}
\usepackage{latexsym}

\usepackage[T1]{fontenc}

\usepackage[utf8]{inputenc}

\usepackage{microtype}

\usepackage{inconsolata}

\usepackage{graphicx}
\usepackage{amsmath}
\usepackage{booktabs}
\usepackage{tabularx}
\usepackage{arydshln} 
\usepackage{amssymb}
\usepackage{longtable}
\usepackage{ltablex}
\usepackage{tcolorbox}
\usepackage{changepage}
\usepackage{verbatim}

\usepackage[normalem]{ulem}
\useunder{\uline}{\ul}{}

\def\method{KAHAN}

\title{\method{}: Knowledge-Augmented Hierarchical Analysis and Narration \\ for Financial Data Narration}

\author{
    Yajing Yang$^{1,2}$, Tony Deng$^{2}$, Min-Yen Kan$^{1}$
    \vspace{2mm} \\
    \textsuperscript{1}National University of Singapore\;\;  \textsuperscript{2}Rio Tinto\\
    {\texttt{\small yajing.yang@u.nus.edu;\;tony.deng@riotinto.com;\;kanmy@comp.nus.edu.sg}}
}

\begin{document}
\maketitle
\begin{abstract}
We propose \method{}, a knowledge-augmented hierarchical framework that systematically extracts insights from raw tabular data at entity, pairwise, group, and system levels. \method{} uniquely  leverages LLMs as domain experts to drive the analysis. On DataTales financial reporting benchmark, \method{} outperforms existing approaches by over 20\% on narrative quality (GPT-4o), maintains 98.2\% factuality, and demonstrates practical utility in human evaluation.
Our results reveal that knowledge quality drives model performance through distillation, hierarchical analysis benefits vary with market complexity, and the framework transfers effectively to healthcare domains. The data and code are available at \url{https://github.com/yajingyang/kahan}.

\end{abstract}

\section{Introduction}

\begin{figure}[t]
    \centering
    \includegraphics[width=\columnwidth]{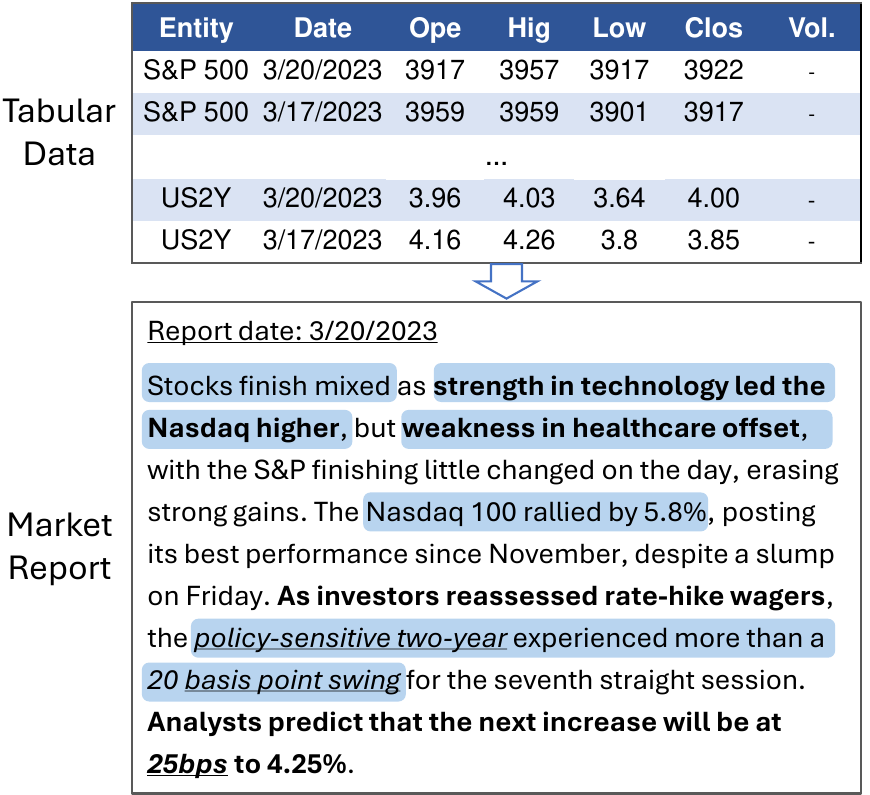}
    \caption{Financial data narration example showing tabular market data (top) and a corresponding market report (bottom). Highlighted phrases of \textbf{domain knowledge analysis}, \textit{\uline{domain-specific terminology}}, and 
    {\setlength{\fboxsep}{1pt}%
\colorbox{cyan!40}{hierarchical relationship analysis}} illustrate how structured narration integrates financial expertise across \texttt{entity} (Nasdaq 100), \texttt{pairwise} (technology vs. Nasdaq), and \texttt{group} (stocks) analytical levels to transform raw market data into contextually meaningful insights.}
    \label{fig:data-narration-example}
    \vspace{-0.5cm}  
\end{figure}

Data narration systems that convert structured data into natural language insights enable automated data interpretation~\cite{dohan_importance_2016, dourish_datafication_2018}. This capability is particularly valuable in financial markets, where analysts transform complex market data into narrative reports that interpret trends, compare performances, and contextualize movements for investment decision-making~\cite{johnson_narrative_2017, michelon_narrative_2022}. While these financial narratives deliver critical value to stakeholders, their manual creation requires significant expertise and time, creating opportunities for automated approaches that bridge raw data and actionable insights.

Data narration systems face two key challenges. First, effective narratives require multi-level analysis---extracting insights at varying granularities and establishing connections between them. In financial contexts, this means analyzing individual assets, sector performance, and market-wide patterns simultaneously. Second, meaningful narration demands domain knowledge augmentation. Recognizing phenomena like tech stock declines during rising interest rates reflect sector-specific capital cost sensitivities requires contextual expertise beyond pattern recognition (Figure~\ref{fig:data-narration-example}).

These challenges expose limitations in current approaches. End-to-end methods flatten data inputs without hierarchical processing and lack domain-specific contextual interpretation~\cite{Puduppully_Dong_Lapata_2019, gong_enhancing_2020, chen_neural_2021}. Large Language Models (LLMs), despite reasoning capabilities and parametric knowledge~\cite{openai2024gpt4, anthropic2024claude, team_gemini_2024}, fail to systematically extract multi-level insights or consistently augment narratives with relevant domain expertise~\cite{chen_knowledge-augmented_2024, zhang_mdsf_2025}. 

To address these limitations, we propose \method{}, a \textbf{K}nowledge-\textbf{A}ugmented \textbf{H}ierarchical  \textbf{A}nalysis and  \textbf{N}arration framework that systematically extracts insights and leverages LLMs as domain experts. Our approach consists of: (1) \textbf{entity-level analysis}  using domain-specific questions for individual insight extraction from raw tabular data, (2) \textbf{multi-level synthesis} generating pairwise, group-level, and system-wide insights through structured analysis, and (3) \textbf{narrative generation} transforming these hierarchical insights into coherent text. Unlike existing approaches that use LLMs merely as text generators, \method{} harnesses their financial reasoning capabilities---demonstrated by performance on specialized assessments like CFA exams~\cite{callanan_can_2023}---to create and apply domain-specific knowledge through structured analytical processes and narrative generation.

Evaluation on the DataTales benchmark~\cite{yang2024datatales} shows that \method{} outperforms existing approaches in providing comprehensive market analysis and actionable investment insights while maintaining factual accuracy, and demonstrating practical utility validated by financial experts. Cross-domain evaluation on healthcare data confirms that \method{}'s approach generalizes to other data narration tasks requiring domain expertise and multi-level analysis.

\section{Related Work}

Data narration has evolved from  template-based approaches~\cite{Reiter_2007, Wiseman_Shieber_Rush_2017} to neural encoder-decoder architectures~\cite{Lebret_Grangier_Auli_2016, Puduppully_Dong_Lapata_2019} that improved fluency but lacked analytical capabilities. While recent LLMs  advance narration through 
improved coherence and reasoning~\cite{openai2024gpt4, anthropic2024claude, team_gemini_2024},
they face two limitations: 
(1) processing data as flattened inputs, lacking the multi-level methodical analysis needed for comprehensive insight extraction~\cite{chen_knowledge-augmented_2024, islam_datanarrative_2024}; 
and (2) inconsistently applying inherent domain knowledge, producing narratives that may miss contextual interpretations~\cite{zhang_mdsf_2025, sui_table_2024}.
These limitations underscore the need for structured frameworks that explicitly guide LLMs in both hierarchical analysis and domain knowledge application.

\paragraph{1. Table Insight Extraction.} 
Table insight extraction has progressed from basic statistical measures~\cite{Sarawagi_Agrawal_Megiddo_1998, Vartak_Rahman_Madden_Parameswaran_Polyzotis_2015} to sophisticated pattern discovery techniques~\cite{Wongsuphasawat_Moritz_Anand_Mackinlay_Howe_Heer_2016, Tang_Han_Yiu_Ding_Zhang_2017, Luo_Qin_Tang_Li_2018, Ma2021MetaInsight}. However, two challenges persist: single-level analysis that misses multi-level relationships~\cite{Demiralp_Haas_Parthasarathy_Pedapati_2017, Srinivasan_Drucker_Endert_Stasko_2019, Zhao2023investigating, Kempf2023KIETA}; and insufficient domain understanding for contextualizing findings~\cite{Law_Endert_Stasko_2020, Ma2023XInsight, Sahu2024InsightBench}. Recent research addresses these gaps through hierarchical analysis systems~\cite{li_coinsight_2024} and domain knowledge augmentation~\cite{he_leveraging_2025}, though none combine these approaches for comprehensive tabular data narration.

\paragraph{2. Knowledge Augmentation.} 
Knowledge augmentation traditionally relied on static knowledge bases~\cite{petroni2019language} and rule systems~\cite{nakano2021webgpt}, evolving to retrieval-augmented generation~\cite{roberts2020knowledge} and knowledge graphs~\cite{chen_knowledge-augmented_2024}. These methods suffer from
post-processing application of knowledge that creates potential misalignments between statistical patterns and domain interpretations, and dependence on labor-intensive expert curation.
Despite LLMs possessing substantial implicit expertise~\cite{bommasani2021opportunities, brown2020language}, current applications underutilize this potential, using them as mere text generators~\cite{kiciman2024causal} or with simplistic prompting~\cite{wei2022chain, kojima2022large}. Even newer techniques exploring automated knowledge generation~\cite{ma_demonstration_2023} and modular adapters~\cite{song_injecting_2025} fail to develop integrated systems where LLM-derived domain knowledge systematically guides the insight discovery process.

\paragraph{}
To overcome these limitations while minimizing manual curation effort, we introduce \method{}, a framework that systematically integrates domain knowledge with hierarchical insight extraction for comprehensive tabular data narration.

\section{\method{} Framework Structure}

\begin{figure}
    \centering
    \includegraphics[width=\columnwidth]{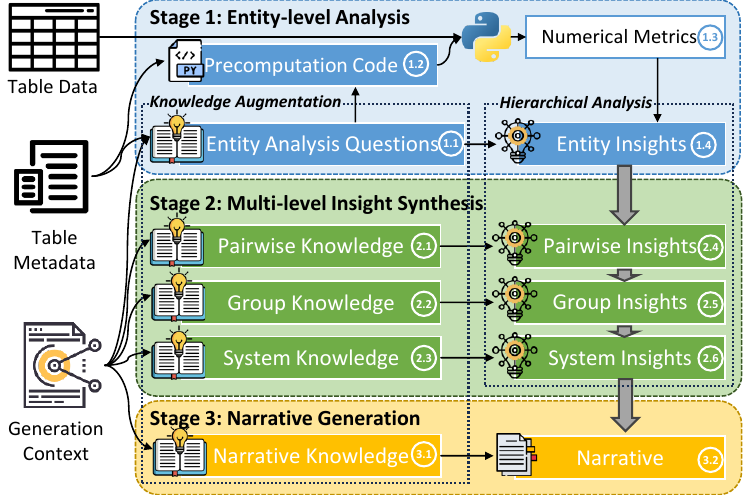}
    \caption{\method{}'s three stages: (1) Entity-level Analysis with question generation, code creation and execution, and insight extraction; (2) Multi-level Insight Synthesis progressing from pairwise, to group, to system-level analysis; and (3) Narrative Generation, producing domain-specific reports.}
    \label{fig:model-structure}
    \vspace{-0.4cm}  
\end{figure}

\begin{figure*}
    \centering
    \includegraphics[width=0.95\textwidth]{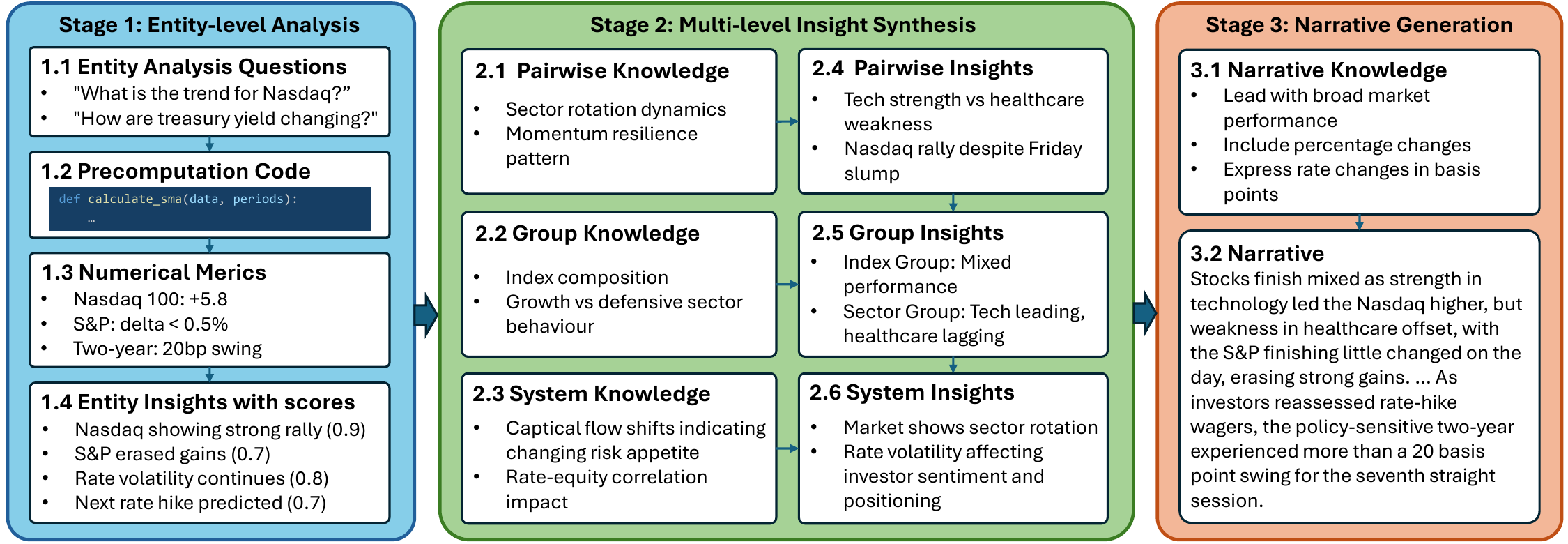}
    \caption{Application of \method{} to financial market data demonstrating hierarchical insight extraction from entity-level observations (individual stock/index performance) through relationship analysis (sector contrasts), group synthesis (Index and Sector groupings), to system-level patterns (market sector rotation), culminating in the structured narrative report shown in Figure~\ref{fig:data-narration-example}.}
    \label{fig:generation-example}
    \vspace{-0.5cm}  
\end{figure*}

\method{} extracts insights from raw tabular data through domain knowledge-guided hierarchical analysis---from single entities, to pairs and groups, to system-level---to output coherent narratives.  We detail this three-stage process (Figure~\ref{fig:model-structure}) with reference to the running example 
in Figure~\ref{fig:generation-example}, which generates the report in Figure~\ref{fig:data-narration-example}.

\subsection{Entity-level Analysis}

The first entity-level analysis stage provides the foundation for hierarchical insight extraction through a four-step process. First, \method{} leverages LLMs to generate domain-specific analytical questions (Step~1.1 in Figure~\ref{fig:model-structure}).  These serve as the semantic foundation for subsequent analysis. This question-driven approach transforms generic data exploration into domain-contextualized analysis, such as momentum metrics in finance or seasonality patterns in retail.

This drives the generation of executable code (Step~1.2) to compute metrics that address the questions. The execution (Step~1.3) of dynamic code enables targeted analytical coverage and domain adaptation, such as customized moving averages for financial trend analysis or weighted inventory turnover in retail contexts.

The final Step~1.4 interprets the numerical results through the analytical questions, producing entity-level insights with significance scores. These data-driven observations (e.g., trend reversal, volatility spikes) provide factual interpretations~\cite{cheng_is_2023} with comprehensive coverage---avoiding the hallucinations and analytical gaps common in direct generation approaches---while serving as building blocks for multi-level synthesis.

\subsection{Multi-level Insight Synthesis}

\method{} synthesizes insights across increasing levels of abstraction, augmenting with domain knowledge at each step to transform entity-level observations into comprehensive dataset understanding.

Each analysis level begins with guiding questions and corresponding knowledge bases: pairwise insights like sector rotation dynamics for entity comparisons; group-level concepts such as index composition structures; and system-wide principles including capital flow indicators of risk appetite (Steps~2.1--2.3). This enables domain-appropriate interpretation beyond generic statistical measures. 


Pairwise analysis (Step~2.4) identifies entity relationships using the generated knowledge base as its analytical framework, comparing high-significance entities to detect meaningful contrast and correction, producing comparative insights like ``Tech strength vs. healthcare weakness'' that quantify specific inter-sector performance dynamics.

Group analysis (Step~2.5) clusters entities into conceptually-related groups using entity classes defined by group-level knowledge, then analyzes aggregate patterns within each group. 
This synthesizes entity-level and pairwise insights to identify group patterns—--revealing cross-market rotation dynamics through ``Index Group'' and ``Sector Group''  identification that are invisible to single-entity analysis.

System-level analysis (Step~2.6) synthesizes all previous insights to identify dataset-wide patterns using systemic indicators from the knowledge base. This holistic synthesis produces macro-level observations that capture complex market dynamics, such as ``Market shows sector rotation'' contextualizing the entire dataset within broader economic frameworks like monetary policy environments.


\subsection{Narrative Generation}



The final stage transforms hierarchical insights into coherent narratives using domain-appropriate structures and language. Narrative knowledge focuses on communicative requirements rather than data interpretation---financial reports require specific section ordering, regulatory disclosures, and audience-appropriate terminology~\cite{michelon_narrative_2022} (Step~3.1), aspects distinct from the analytical frameworks used in earlier stages.

The generation algorithm (Step~3.2) produces narratives that balance entity-level details with relationship patterns and system-level observations, creating a coherent flow rather than disconnected sections. 

\paragraph{}A key advantage of \method{} is knowledge reusability across all three stages: domain knowledge bases can be cached for subsequent reports with minimal adaptation, reducing computation for similar data contexts (e.g., daily market reports for different dates) while maintaining consistent analytical frameworks. This enables efficient processing while ensuring entity-specific insights still reflect current data patterns.

\section{Evaluation}

We evaluate \method{} against baseline approaches using various models and evaluation settings.

\begin{table*}[t]
\centering
\resizebox{0.8\textwidth}{!}{%
\begin{tabular}{@{}clccccc@{}}
\toprule
 &  & \multicolumn{4}{c}{Quality} & Factuality \\ \midrule
Model & Setting & Agg. Score & Description & Insights & Readability & FActScore (\%) \\ \midrule
 & w/ DP & 6.48 & 6.49 & 6.40 & 6.62 & 96.7 \\
Llama3.1-8B-Instruct & w/ CoT & 5.27 & 5.30 & 5.17 & 5.39 & \textbf{99.6} \\
 & w/ KAHAN & \textbf{6.97} & \textbf{7.06} & \textbf{6.96} & \textbf{6.80} & 97.8 \\ \midrule
 & w/ DP & 7.11 & 7.12 & 7.03 & 7.23 & 97.7 \\
Qwen2.5-7B-Instruct & w/ CoT & 6.43 & 6.53 & 6.33 & 6.46 & \textbf{98.4} \\
 & w/ KAHAN & \textbf{7.86} & \textbf{8.07} & \textbf{7.81} & \textbf{7.56} & 94.1 \\ \midrule
 & w/ DP & 6.89 & 6.92 & 6.59 & \textbf{7.42} & \textbf{99.7} \\
GPT-4o & w/ CoT & 6.61 & 6.71 & 6.38 & 6.90 & 99.1 \\
 & w/ KAHAN & \textbf{8.26} & \textbf{8.58} & \textbf{8.39} & 7.34 & 98.2 \\ \bottomrule
\end{tabular}
}
\caption{Quality (0--10 scale) and factuality (percentage) comparison across models and settings. For each model, the highest values per column are \textbf{bolded}. \method{} consistently outperforms Direct Prompting (DP) and Chain-of-Thought (CoT) on aggregate quality score, description quality, and insight generation.}
\label{tab:main-results}
\vspace{-0.4cm}  
\end{table*}



\paragraph{Models.} We test our framework using open source models \textit{Llama3.1-8B-instruct}~\cite{grattafiori_llama_2024} (temperature=0.7, top\_p=0.8) and \textit{Qwen2.5-7B-instruct}~\cite{qwen_qwen25_2025} (temperature=0.7, top\_p=0.8, repetition\_penalty=1.05), along with  proprietary model \textit{GPT-4o}~\cite{openai_gpt-4o_2024} (temperature=1.0), spanning different scales and types. We also tried to compare using LLMs pretrained specifically on financial corpora---FinanceLLM~\cite{cheng2024adapting} and TouchStoneGPT~\cite{wu_golden_2024}---but both failed at Stage~1 (code generation) and Stage~2 (structured analysis output), hence were omitted.



\paragraph{Experimental Setup.} We compare \method{} against two baselines: \textit{Direct Prompting (DP)} and \textit{Chain of Thought (CoT)}. DP implements zero-shot generation of entity-level insights and narratives without explicit domain knowledge or intermediate analysis steps; CoT enhances DP by incorporating step-by-step reasoning instructions in each stage (details in Appendix~\ref{sec:appendix-generation-prompt}).


\paragraph{Evaluation Metrics.} We evaluate our approach on the \textit{DataTales}~\cite{yang2024datatales} benchmark testset, which contains 460 samples spanning 11 financial markets with varying numbers of entities per market. We assess the generated narratives on quality, factuality, and practical utility.

For \textbf{quality}, we adopt DnA-Eval~\cite{li_dna-eval_2024} an automated, multi-dimensional evaluation via aspect-specific scoring rather than holistic judgment.
To implement this approach, we used GPT-4o to identify the most critical aspects for financial narrative evaluation, resulting in three key dimensions with corresponding weights (in parentheses) for the aggregated quality score: 
\begin{itemize}
    \item \textbf{Description (40\%)}: ``Does the output provide clear and comprehensive data on current market trends and conditions?''
    \item \textbf{Insights (40\%)}: ``Does the output include actionable insights and analysis to support investment decision-making?''
    \item \textbf{Readability (20\%)}: ``Is the information in the output presented in an easily understandable and accessible format?''
\end{itemize}
We evaluate all methods' generations for the same input together, assigning aspect-specific scores (0-10), and aggregated for a final quality score. We repeat the evaluation for 3 times and report the average for aspect-specific and aggregated quality scores. All evaluations use temperature 0.1 to reduce inconsistency.
These dimensions align with quality criteria established in financial reporting literature~\cite{huang_evidence_2014, michelon_narrative_2022}, ensuring our assessment reflects professional standards.

For \textbf{factuality}, we implement a modified version of FActScore~\cite{min_factscore_2023} that decomposes generated text into atomic facts and verifies each against reliable knowledge sources. Our implementation replace the knowledge base setup in FActScore (Wikipedia articles) in two ways: first, using manually coded scripts that cover 96\% of the metrics calculated in Step~3.3 across all generations; and second, incorporating information extracted from Wikipedia's finance category using vector search to verify domain-specific claims. Both atomic fact decomposition and knowledge-based validation utilize GPT-4o mini. 

For \textbf{practical utility}, we conduct human evaluation with 2 financial domain experts: one trader (8 years experience) and one analyst (3 years experience). Experts blindly rank outputs from \method{}, CoT, and DP with GPT-4o on 30 financial narratives based on "usefulness for investment decision-making." The trader represents our target audience of investors requiring actionable market insights.

We include the detailed prompts for both quality (including aspect generation) and factuality (including numerical metrics) evaluation in Appendix~\ref{sec:appendix-evaluation-setup} for replicability.

\section{Results and Analyses}


Table~\ref{tab:main-results} shows that \method{} consistently outperforms both DP and CoT approaches, with largest gains in quality metrics (Description and Insights). With GPT-4o, \method{} achieves a quality score of 8.26, representing a 20\% improvement over DP (6.89) and 25\% over CoT (6.61).  This trend holds across model scales while maintaining remarkably high factuality for Llama3.1 and GPT-4o (97.8-98.2\%), with only Qwen2.5 showing a slight trade-off (94.1\%). \method{} also demonstrates superior stability across quality evaluation runs (detailed in Appendix~\ref{sec:appendix-variance-analysis}). For each setting, we sample 30 generations to analyze coverage metrics (cf Table~\ref{tab:quantitative-analysis}). 
Ablation studies isolate contributions by removing individual knowledge components (entity insights, insight synthesis, narrative processing) (Figure~\ref{fig:knowledge-ablation}) and evaluating progressive hierarchical complexity (entity-only $\rightarrow$  entity+pairwise $\rightarrow$  entity+pairwise+group $\rightarrow$  full KAHAN) (Figure~\ref{fig:hierarchical-ablation}).

\subsection{Descriptive Quality}

\method{} demonstrates substantial improvements in descriptive quality across all models (15\% over DP, 29\% over CoT). This enhancement stems from three key factors: broader market coverage (\#1, 4.59--5.17 vs. 1.83--3.11 indicators in baselines), enhanced contextual awareness (\#4, 45--85\% inclusion of macroeconomic context vs. 25--64\%), and more comprehensive timeframe analysis (\#3).

Comparing generation snippets (full reports in Appendix~\ref{sec:appendix-generation_example_baseline}) illustrates these deltas:

\begin{tcolorbox}[
    colback=white,
    colframe=black,
    boxrule=0.5pt,
    arc=0pt,
    outer arc=0pt,
    left=5pt,
    right=5pt,
    top=5pt,
    bottom=5pt,
    boxsep=0pt,
    fontupper=\small
]
\textbf{\method{}}: ``The S\&P 500 closed at 4016.95, showcasing positive short-term and medium-term momentum by remaining above its 10-day and 50-day simple moving averages (SMAs). Despite this, the index experienced reduced trading volume and lower-than-average intraday volatility, suggesting a cautious market sentiment.''\\

\textbf{DP}: ``The S\&P 500 closed at 4016.95, significantly above its 20-day Simple Moving Average (SMA) of 3905.881.''
\end{tcolorbox}

\begin{figure}
    \centering
    \includegraphics[width=0.9\columnwidth]{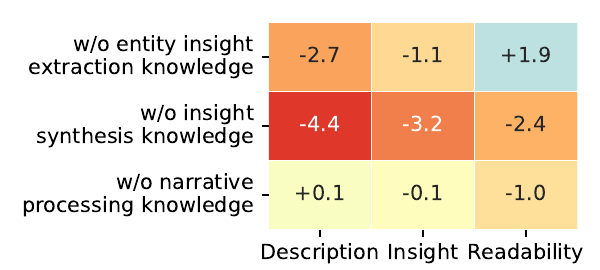}
    \caption{Knowledge components ablation in Llama-3.1, showing percentage change in quality aspects when each component is ablated.} 
    \label{fig:knowledge-ablation}
\end{figure}

Ablation studies reveal how our framework achieves these improvements. First, domain-specific knowledge significantly impacts description quality, with entity insight extraction ($-$0.19 points) enabling the broader technical indicator coverage seen in our example (multiple SMAs, volume, volatility), while insight synthesis ($-$0.31 points) facilitates the contextual awareness that connects these indicators to market sentiment (Figure~\ref{fig:knowledge-ablation}). Second, hierarchical structure benefits vary by model capability—Llama3.1 gains most benefits at the entity-pairwise level, while GPT-4o shows pronounced improvement (12.1\%) with the complete hierarchical structure (Figure~\ref{fig:hierarchical-ablation}), suggesting more powerful models better leverage complex hierarchical analysis for enhanced description.

\begin{table}[]
\resizebox{\columnwidth}{!}{%
\begin{tabular}{@{}ccccc@{}}
\toprule
\textbf{Quantitative Metrics} & \textbf{Settings} & \textbf{Llama3.1} & \textbf{Qwen2.5} & \multicolumn{1}{l}{\textbf{GPT-4o}} \\ \midrule
 & DP & 2.73 & 2.10 & 1.83 \\
\textbf{1. Market indicators} & CoT & 2.00 & 2.48 & 3.11 \\
 & KAHAN & 4.87 & 4.59 & 5.17 \\ \midrule
 & DP & 4.47 & 4.30 & 4.03 \\
\textbf{2. Sectors/asset classes} & CoT & 3.27 & 4.00 & 3.43 \\
 & KAHAN & 4.23 & 4.10 & 4.73 \\ \midrule
 & DP & 1.93 & 1.40 & 1.23 \\
\textbf{3. Timeframe considerations} & CoT & 1.47 & 1.38 & 1.57 \\
 & KAHAN & 1.80 & 1.76 & 1.73 \\ \midrule
 & DP & 30\% & 25\% & 42\% \\
\textbf{4. Macroeconomic context (\%)} & CoT & 38\% & 28\% & 64\% \\
 & KAHAN & 85\% & 45\% & 72\% \\ \midrule
 & DP & 80\% & 82\% & 98\% \\
\textbf{5. Comparative data (\%)} & CoT & 42\% & 81\% & 93\% \\
 & KAHAN & 87\% & 100\% & 100\% \\ \midrule
 & DP & 1.23 & 1.77 & 1.53 \\
\textbf{6. Causal relationships} & CoT & 0.80 & 2.10 & 2.18 \\
 & KAHAN & 3.03 & 4.17 & 4.87 \\ \midrule
 & DP & 1.40 & 1.53 & 1.53 \\
\textbf{7. Analysis present} & CoT & 1.07 & 1.79 & 1.86 \\
 & KAHAN & 2.07 & 2.62 & 3.03 \\ \midrule
 & DP & 2.27 & 1.37 & 1.27 \\
\textbf{8. Explicit recommendations} & CoT & 1.27 & 2.10 & 1.50 \\
 & KAHAN & 2.27 & 1.69 & 1.73 \\ \midrule
 & DP & 57\% & 40\% & 52\% \\
\textbf{9. Risk assessment (\%)} & CoT & 47\% & 74\% & 82\% \\
 & KAHAN & 65\% & 93\% & 98\% \\ \midrule
\textbf{} & DP & 2.72 & 3.07 & 3.03 \\
\textbf{10. Structure elements} & CoT & 1.83 & 2.86 & 3.04 \\
\textbf{} & KAHAN & 2.70 & 3.69 & 3.90 \\ \midrule
\textbf{} & DP & 5.00 & 1.48 & 3.19 \\
\textbf{11. Flesch-Kincaid Grade Level} & CoT & 7.00 & 4.96 & 4.29 \\
\textbf{} & KAHAN & 7.05 & 3.74 & 7.50 \\ \bottomrule
\end{tabular}
}
\caption{Quantitative analysis. Counts of key narrative elements (market indicators, sectors, timeframes, causal relationships, analysis levels, recommendations), inclusion rates (macroeconomic context, comparative data, risk assessment), and readability metrics (structure score, Flesch--Kincaid Grade Level (FKGL) for each model--setting combination. Higher values are better, except for FKGL.}
\label{tab:quantitative-analysis}
\vspace{-0.4cm}  
\end{table}

\subsection{Insight Quality}
\method{} delivers its most substantial gains in insight quality across all models (16\% over DP, 26\% over CoT), driven by three critical capabilities: enhanced causal understanding (3.03--4.87 relationships vs. 0.80--2.18 in baselines), multi-level analysis spanning micro-to-macro factors (2.17--2.83 levels of analysis present vs. 1.27--1.93 in baselines), and consistent risk assessment integration (65--98\% vs. 40--82\%).

A peek at a single statement demonstrates all three key enhancements: the \method{} version establishes a causal relationship between bond yields and sector performance (causal understanding), connects macro factors (bond yields) with sector-specific performance (multi-level analysis), and incorporates risk assessment (``bearish momentum'' and ``cautious outlook'').

\begin{tcolorbox}[
colback=white,
colframe=black,
boxrule=0.5pt,
arc=0pt,
outer arc=0pt,
left=5pt,
right=5pt,
top=5pt,
bottom=5pt,
boxsep=0pt,
fontupper=\small
]
\textbf{\method{}}: ``The performance of this sector is influenced by changes in the U.S. 10-Year Bond Yield, which shows slight bearish momentum and negative daily performance, reinforcing a cautious outlook.''\\

\textbf{DP}: ``The S\&P 500 Financials mirrored this trend, closing at 595.37 compared to a 20-day SMA of 583.0345, suggesting positive market momentum.''
\end{tcolorbox}

\begin{figure}
    \centering
    \includegraphics[width=\columnwidth]{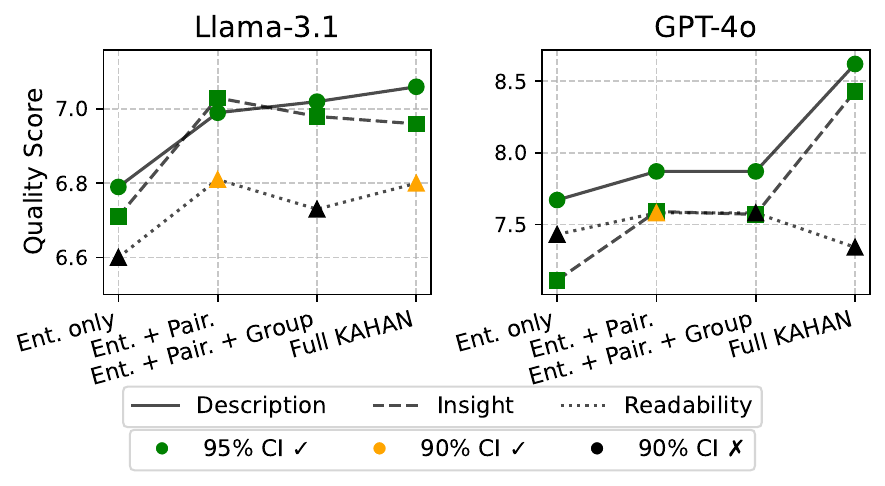}
    \caption{Hierarchical analysis ablation in Llama-3.1 and GPT-4o. The line plots show performance changes as hierarchical components are progressively added from left to right. The x-axis uses "Ent." for entity and "Pair." for pairwise analysis. 
    Marker colors indicate statistical significance of improvements, with CI denoting confidence interval (detailed in Appendix~\ref{sec:appendix-hierarchical-significant-testing})}
    \label{fig:hierarchical-ablation}
    \vspace{-0.4cm}  
\end{figure}

Ablation studies show insight quality depends more on synthesized knowledge ($-0.22$ points when removed) than entity extraction knowledge ($-0.08$), similar to description quality (Figure~\ref{fig:knowledge-ablation}). The hierarchical structure affects insight generation differently across models: Llama3.1 reaches peak insight quality at the entity+pairwise level (7.03) with minimal gains beyond this (Figure~\ref{fig:hierarchical-ablation}). In contrast, GPT-4o shows moderate improvement at pairwise level ($+0.43$) but substantial gains ($+0.84$) with system-wide analysis. This indicates that larger models can effectively leverage complete hierarchical structures to identify complex market interactions and causal relationships.

\subsection{Readability}
\method{} demonstrates an interesting pattern in readability performance, showing consistent improvements over CoT ($+16.5\%$ on average) but model-dependent relationship with DP—better for smaller models (Llama3.1: $+2.7\%$, Qwen2.5: $+4.6\%$) but slightly lower for GPT-4o ($-1.1\%$). This suggests that while structured processing benefits less capable models, advanced models may already excel at information organization.

Analysis of structural elements reveals \method{} produces better-organized outputs (2.70--3.90 structural elements) compared to CoT (1.83--3.04), while linguistic complexity varies based on model capabilities—more sophisticated for Llama3.1 and GPT-4o, but more accessible for Qwen2.5, aligning with readability scores.

Our ablation studies show that removing narrative processing knowledge has minimal impact on readability ($-0.07$) (Figure~\ref{fig:knowledge-ablation}), while readability peaks at the entity+pairwise level for both models (Llama3.1: 6.81, GPT-4o: 7.59) and slightly decreases with additional hierarchical levels (Figure~\ref{fig:hierarchical-ablation}). This suggests that intermediate-level hierarchical structure—--rather than narrative guidance—--optimizes readability, while more comprehensive levels prioritize insights at some cost to accessibility. This may explain why GPT-4o shows less dramatic improvement in this dimension compared to other aspects.

\subsection{Factuality}

\method{} maintains high factuality (94.1--98.2\%) while delivering superior narrative quality. GPT-4o with \method{} achieves both excellent aggregated quality score (8.33) and strong factuality (98.2\%), demonstrating that larger models effectively manage the quality--factuality balance. In contrast, Llama3.1 with \method{} improves both quality and factuality over DP, while approaching CoT's exceptional factuality (97.8\% vs. 99.6\%), while Qwen2.5 prioritizes quality improvements.

Our error analysis categorize factual error by complexity: for GPT-4o, interpretive claims about market sentiment (e.g., ``RSI signals robust upward momentum'') account for 63\% for all errors, while metrics calculations (e.g., ``a --9.07\% change'') represent 28\%, and raw data reporting errors just 9\%, suggesting that factuality accuracy inherently decreases as analytical complexity increases.

\subsection{Human Evaluation of Practical Utility}

Expert evaluation reveals distinct professional preferences that align with \method{}'s design objectives. The trader find \method{} most useful in 80\% of cases, while analysts prefer CoT's conciseness (56.7\% vs. \method{}'s 20\%). This divergence reflects fundamental differences in professional requirements: traders need comprehensive analytical depth for decision-making, while the analyst prioritize brevity for report consolidation.

This strong trader preference validates \method{}'s hierarchical decomposition strategy for its intended audience of investors seeking actionable market analysis. These results also demonstrate that our automated quality metrics successfully predict real-world utility when properly aligned with the target application domain.

\section{Discussion: Scaling Factors and Cross-domain Applicability}

We examine three key factors that influence \method{}'s effectiveness: (1) knowledge quality, comparing performance when domain knowledge is generated using different models and methods, (2) market complexity, assessing how our hierarchical approach performs on markets with varying numbers of entities (few vs. many), and (3) domain transferability, evaluating how our framework generalizes beyond financial data to healthcare analytics. 

\begin{figure}
    \centering
    \includegraphics[width=\columnwidth]{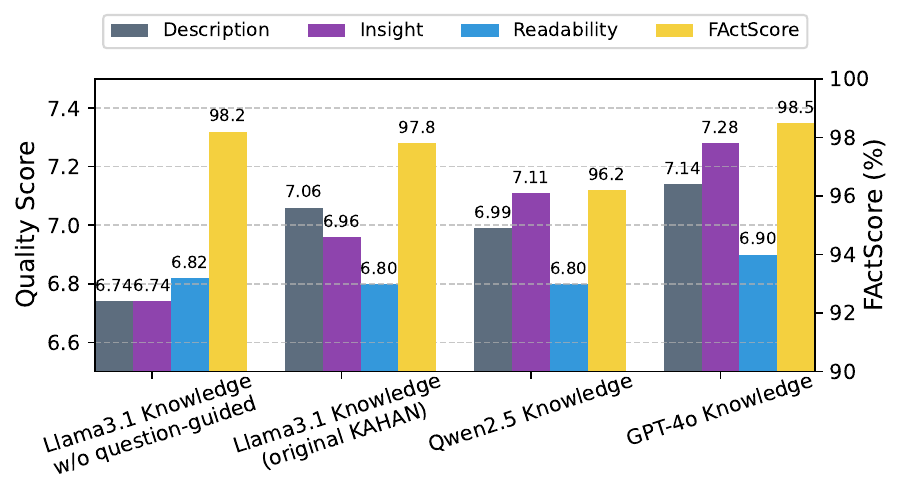}
    \caption{Impact of knowledge source on data narrative quality.}
    \label{fig:knowledge-impact}
    \vspace{-0.4cm}  
\end{figure}

\subsection{RQ1: Impact of Knowledge Quality on Generation Performance}

\textit{How does knowledge source quality affects narrative generation?} We test this by keeping Llama3.1-8B as the base model, while varying knowledge generation methods and sources (Figure~\ref{fig:knowledge-impact}).

Question-guided knowledge generation significantly improves performance compared to unguided generation (+0.21 aggregate points, +0.32 description, +0.22 insight), confirming that structured acquisition creates more effective knowledge bases for market narration.


Regarding knowledge source models, we observe distinct performance patterns that mirror our main results. Qwen2.5-sourced knowledge enhances insight quality but introduces a slight factuality decline compared to the original Llama3.1 knowledge. In contrast, GPT-4o-generated knowledge enables Llama3.1 to achieve the highest performance across all dimensions simultaneously, reinforcing that high-quality knowledge can overcome the quality-factuality trade-off. This demonstrates that smaller models can be performant through knowledge distillation from more powerful models, providing an efficient pathway for high-quality financial narrative systems requiring reduced compute.

\subsection{RQ2: Impact of Market Complexity on Hierarchical Analysis Effectiveness}

\begin{figure}
    \centering
    \includegraphics[width=\columnwidth]{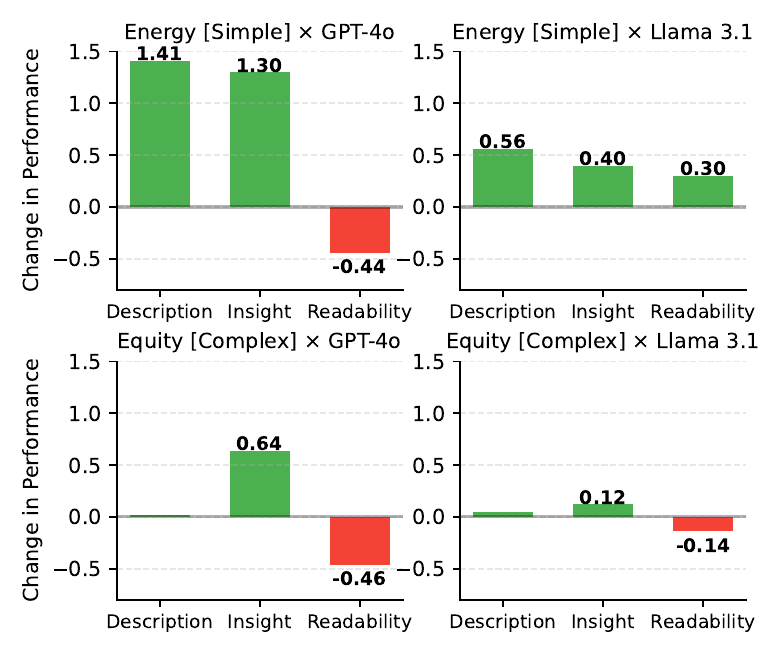}
    \caption{Impact of hierarchical analysis across market complexity on data narrative quality.}
    \label{fig:hierarchical-impact}
    \vspace{-0.4cm}  
\end{figure}

\textit{How does the complexity of the market impact the analysis effectiveness?} 
We test this by varying market complexity using two models (GPT-4o, Llama3.1) and two market types (equity markets with 28+ entities including indices, sectors, and treasury yields; energy markets with 3 entities: WTI, Brent, and Natural Gas). For each combination, we compared 30 sampled generations with and without hierarchical analysis (Figure~\ref{fig:hierarchical-impact}).

Our analysis reveals three significant patterns. First, simpler markets derive substantially greater benefits from hierarchical analysis. With GPT-4o, energy markets show dramatic improvements (description: +1.41, insight: +1.30), compared to modest gains in equity markets (description: +0.02, insight: +0.64). Second, insight quality consistently improves across all conditions, confirming relationship identification remains valuable regardless of market structure. Third, complex markets exhibit a clear insight--readability trade-off absent in simpler markets, where Llama3.1 actually shows improved readability (+0.30).

Generation examples in Appendix~\ref{sec:appendix-market-complexity} highlight these differences: complex market narratives identify sophisticated cross-asset relationships (``inverse relationship between Dollar Index and Gold''), while simple market narratives achieve deeper comparative analysis with fewer entities (``contrasting volatility patterns between Brent Crude Oil and Natural Gas''). These findings suggest optimal system design should adaptively adjust hierarchical depth based on market complexity---applying full hierarchical analysis for simpler markets, while limiting analytical complexity for markets with numerous entities.

\subsection{RQ3: Transferability to Healthcare Domain}

\begin{figure}
    \centering
    \includegraphics[width=\columnwidth]{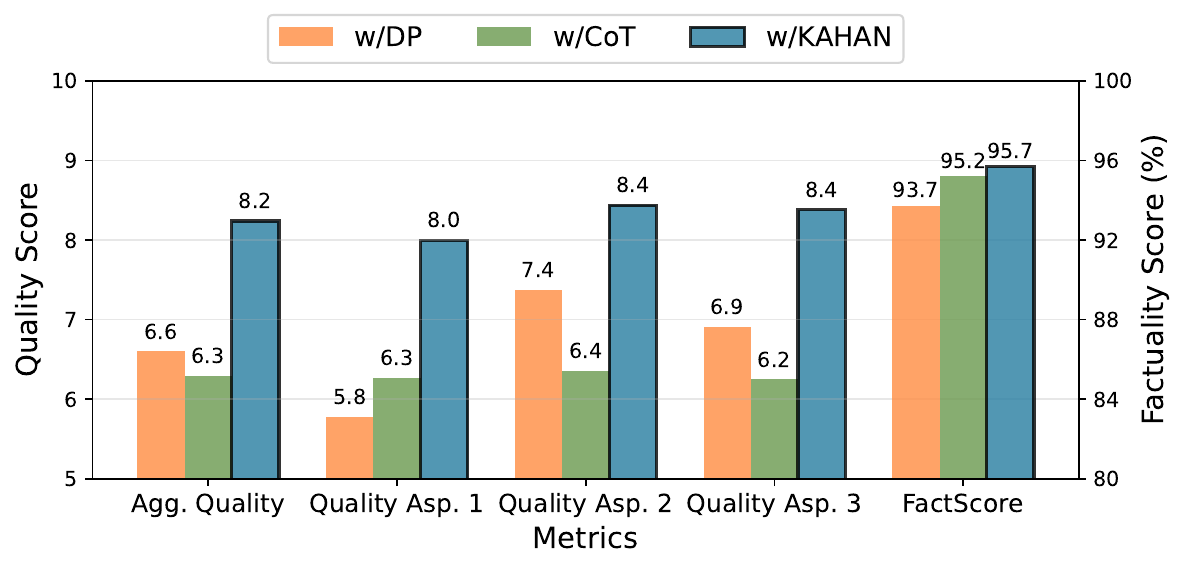}
    \caption{Cross-domain performance on healthcare data.}
    \label{fig:healthcare-application}
    \vspace{-0.4cm}  
\end{figure}
\textit{How effectively does \method{} transfer to non-financial domains with specialized knowledge requirements?}
We apply \method{} to Parkinson's Disease (PD) gait analysis from PhysioNet—a healthcarae dataset of force sensor recordings from 93 PD patients and 73 healthy controls across 306 sessions~\cite{Goldberger2000-ah}. This domain shares key characteristics with finance: specialized knowledge requirements, multi-entity analysis, and decision-support objectives, while differing in expertise domain (medical vs. economic) and audience expertise (knowledgeable investors vs. general patients).


Results with Llama-3.1 show \method{} substantially outperforms baseline approaches across all evaluated dimensions, achieving an aggregate quality score of 8.24 compared to 6.29--6.60 for alternatives while maintaining comparable factual accuracy (95.7\% vs. 93.7--95.2\%) (Figure~\ref{fig:healthcare-application}). This comprehensive improvement demonstrates that \method{}'s integration of domain knowledge with structured analysis successfully adapts to different specialized domains, generating contextually appropriate insights regardless of the underlying subject matter expertise required.

\section{Conclusion}


This paper introduces \method{}, a knowledge-augmented hierarchical framework that transforms financial data into coherent narratives. Our approach outperforms baseline methods while revealing insights about effective narrative systems. Knowledge augmentation enhances narrative quality while maintaining factuality, enables distillation to smaller models, and transfers effectively to other expertise-intensive domains. Hierarchical analysis improves insight discovery and narrative organization, with benefits varying by market complexity and practical utility validated by financial experts.

Future research directions include: extending knowledge augmentation by incorporating real-time market news~\cite{puh_predicting_2023}, developing multi-modal frameworks with visualizations to balance readability and analysis depth~\cite{chy_role_2023}, integrating predictive analytics for actionable insights based on identified causal relationships~\cite{he_financial_2023}.


\section*{Acknowledgments}
We thank Teoh Jun Yeang and Gong Xue for their valuable time and expertise in conducting the human evaluation of practical utility for financial narrative generation.

\section*{Limitations}

Our evaluation of \method{} currently focuses exclusively on financial market data due to time and resource constraints, despite its design as a general data narration framework. Testing across diverse domains remains necessary to verify its broad applicability. Additionally, our evaluation methodology relies on expert assessment of narrative quality rather than measuring downstream impact on decision-making outcomes. Future work should examine how narratives generated by \method{} influence actual investment decisions and performance compared to human-generated alternatives

\section*{Ethical Considerations}

While \method{} is designed to augment rather than replace financial analysts, it may raise concerns about overreliance on automated narratives and potential reinforcement of existing biases in financial interpretation. Our hierarchical approach with explicit step-by-step reasoning helps mitigate these risks by making the analysis process transparent and auditable. However, the quality of insights remains dependent on the underlying models' domain knowledge, which may contain outdated or incomplete financial concepts. Additionally, while we've focused on improving factuality, sophisticated users could potentially misuse the system to create misleading yet plausible-sounding market narratives that appear authoritative. Future work should explore safeguards against such deceptive applications.

\bibliography{custom}

\appendix
\newpage

\section{Generation Prompts}
\label{sec:appendix-generation-prompt}
\subsection{\method{}}

\subsubsection{Stage 1: Entity-level Analysis}
Prompts to generate entity data analytical questions:

\begin{tcolorbox}[
colback=white,
colframe=black,
boxrule=0.5pt,
arc=0pt,
outer arc=0pt,
left=5pt,
right=5pt,
top=5pt,
bottom=5pt,
boxsep=0pt,
fontupper=\small,
]
You are a \{domain\} expert focused on \{market\}.\\
Your eventual goal is to \{task\}.\\
For now, generate 5 comprehensive analysis questions to extract insights for the task.\\

Available data fields:\\
\{table\_schema\}\\

Task:
Generate analysis questions that:\\
1. Cover different aspects of entity analysis\\
2. Can be answered using quantifiable metrics\\
3. Provide meaningful insights for \{task\}\\

For each question:\\
1. Specify a clear insight type\\
2. List ALL metrics that need to be pre-computed\\
3. Consider relative comparisons if relevant\\

Structure output as JSON list:
\begin{verbatim}
[
  {{
    "insight_type": "descriptive name of insight 
    type",
    "question": "analysis question with ALL 
    required metrics andcalculations specified",
    "required_metrics": [
      {{
        "metric": "metric name",
        "calculation": "how to compute",
        "purpose": "how used in analysis"
      }}
    ],
    "comparisons": ["relevant comparisons"]
  }}
]
\end{verbatim}

Example output:
\begin{verbatim}
[
  {{
    "insight_type": "trend",
    "question": "On the date of interest, 
    analyze the trend direction and strength 
    by comparing price 
    against multiple technical indicators, 
    including moving averages, momentum indica-
    tors, and support/resistance levels",
    "required_metrics": [
      {{
        "metric": "SMA_20",
        "calculation": "20-day Simple Moving 
        Average of closing_price",
        "purpose": "Identify short-term trend 
        direction"
      }},
      {{
        "metric": "MACD_line",
        "calculation": "12-day EMA minus 26-day 
        EMA of closing_price",
        "purpose": "Measure trend momentum and 
\end{verbatim}
\end{tcolorbox}

\begin{tcolorbox}[
colback=white,
colframe=black,
boxrule=0.5pt,
arc=0pt,
outer arc=0pt,
left=5pt,
right=5pt,
top=5pt,
bottom=5pt,
boxsep=0pt,
fontupper=\small
]
\begin{verbatim}
        possible reversals"
      }},
      {{
        "metric": "support_level",
        "calculation": "Minimum price levels with 
        multiple bounces in last 30 days",
        "purpose": "Identify price support areas"
      }},
      ...
    ],
    "comparisons": [
      "Current price vs moving averages",
      "MACD line vs signal line",
      "Price position relative to support/
      resistance",
      "Current vs historical trend strength"
    ]
  }},
  ...
]
\end{verbatim}

Ensure the questions:\\
1. Are specific and quantifiable\\
2. Use available data fields\\
3. Cover different analytical aspects\\
4. Support {self.task} objectives\\
5. Are appropriate for {self.market} analysis\\
6. Assigned with a different insight type\\

Return only the JSON object by presenting the list directly without meta-commentary, introductions, or language specification (i.e., start with "[") and excluding concluding remarks or follow-up suggestions (i.e., ends with "]")
\end{tcolorbox}

Prompt to generate metrics computation code:
\begin{tcolorbox}[
colback=white,
colframe=black,
boxrule=0.5pt,
arc=0pt,
outer arc=0pt,
left=5pt,
right=5pt,
top=5pt,
bottom=5pt,
boxsep=0pt,
fontupper=\small
]
Write Python code to perform the required calculations and save results.

Question: {question}

Requirements:

1. Read data into pandas DataFrame from argparser argument --data\_path with type str\\
2. The entity name is provided as argparser argument --product\_name with type str\\
3. The date of interest is provided as argparser argument --date with type str\\
4. Process the data to the data types specified below, be careful with the data types during operation\\
5. Handle missing/invalid values in the data\\
6. For time-series calculations (e.g., moving averages):
\begin{adjustwidth}{0.4cm}{0cm}
- Maintain complete historical data until final calculation\\
- Only filter by date after computing time-dependent metrics\\
- Ensure proper handling of lookback periods
\end{adjustwidth}
7. Print results (e.g., sma-20) and values (closing price) required in a structured format:\\
For each item:
\begin{adjustwidth}{0.4cm}{0cm}
- Print "METRIC:" followed by the metric name\\
- Print "VALUE:" followed by the calculated value\\
- Print "UNIT:" followed by the unit (if applicable)\\
- Print "TYPE:" followed by the data type of the result\\
- Example: METRIC:monthly\_return VALUE:0.0234 UNIT:percent TYPE:float
\end{adjustwidth}
8. Use only the following columns from data:
\begin{adjustwidth}{0.4cm}{0cm}
\{table\_schema\}\\
\end{adjustwidth}

Return the only the python code without meta-commentary, introductions, or language specification
\end{tcolorbox}

\subsubsection{Stage 2: Multi-level Insight Synthesis}
Prompt to generate pairwise knowledge:

\begin{tcolorbox}[
colback=white,
colframe=black,
boxrule=0.5pt,
arc=0pt,
outer arc=0pt,
left=5pt,
right=5pt,
top=5pt,
bottom=5pt,
boxsep=0pt,
fontupper=\small
]
You are a \{domain\} expert focused on \{market\}. 
Your eventual goal is to \{task\}.
For now, generate domain knowledge about relationships and patterns based on these analysis questions.

Context:
Entities in scope: \{entities\}

Insight types being analyzed:\\
\{insight\}\\
\{pairwise\_questions\}\\
Task:\\
For each category of questions, provide domain knowledge about:\\
1. Typical patterns and relationships\\
2. Common influencing factors\\
3. Important conditions and contexts\\
4. Notable exceptions or special cases\\

Structure output as JSON:
\begin{verbatim}
{{
  "<knowledge_group_name>": [
    {{
      "key_idea": <analysis question>,
      "description":  <answer to analysis question>
    }},
    ...
  ],
  ...
}}
\end{verbatim}

Provide knowledge that is:\\
- Specific to \{domain\}\\
- Relevant for analyzing the given entities and insight types\\
- Based on established patterns and relationships\\
- Applicable across different scenarios\\
- Clear and actionable for analysis\\

Return only the JSON object by presenting the list directly without meta-commentary, introductions, or language specification (i.e., start with "\{\{") and excluding concluding remarks or follow-up suggestions (i.e., ends with "\}\}") 
\end{tcolorbox}

prompt to generate pairwise knowledge:

\begin{tcolorbox}[
colback=white,
colframe=black,
boxrule=0.5pt,
arc=0pt,
outer arc=0pt,
left=5pt,
right=5pt,
top=5pt,
bottom=5pt,
boxsep=0pt,
fontupper=\small
]
You are a \{domain\} expert focused on \{market\}. \\
Your eventual goal is to \{task\}.\\
For now, generate domain knowledge about relationships and patterns based on these analysis questions.\\

Context:\\
Entities in scope: \{entities\}\\
Insight types being analyzed:
\{insight\}\\
\{pairwise\_questions\}\\

Task:\\
For each category of questions, provide domain knowledge about:\\
1. Typical patterns and relationships\\
2. Common influencing factors\\
3. Important conditions and contexts\\
4. Notable exceptions or special cases

Structure output as JSON:
\begin{verbatim}
{{
  "<knowledge_group_name>": [
    {{
      "key_idea": <analysis question>,
\end{verbatim}
\end{tcolorbox}

\begin{tcolorbox}[
colback=white,
colframe=black,
boxrule=0.5pt,
arc=0pt,
outer arc=0pt,
left=5pt,
right=5pt,
top=5pt,
bottom=5pt,
boxsep=0pt,
fontupper=\small
]
\begin{verbatim}
      "description":  <answer to analysis question>
    }},
    ...
  ],
  ...
}}
\end{verbatim}
Provide knowledge that is:\\
- Specific to {domain}\\
- Relevant for analyzing the given entities and insight types\\
- Based on established patterns and relationships\\
- Applicable across different scenarios\\
- Clear and actionable for analysis\\

Return only the JSON object by presenting the list directly without meta-commentary, introductions, or language specification (i.e., start with "\{\{") and excluding concluding remarks or follow-up suggestions (i.e., ends with "\}\}") 
\end{tcolorbox}

Prompt for pairwise synthesis to analyze high significant entity insights with other entities in the dataset:
\begin{tcolorbox}[
colback=white,
colframe=black,
boxrule=0.5pt,
arc=0pt,
outer arc=0pt,
left=5pt,
right=5pt,
top=5pt,
bottom=5pt,
boxsep=0pt,
fontupper=\small
]
As a \{domain\} expert targeting to \{task\}, analyze insights for this entity cluster using domain knowledge.\\

Cluster: \{cluster\_name\}\\
Member Entities: \{cluster\_entities\}\\

Entity Insights: \{cluster\_entity\_insights\}\\
Pairwise Relationship Insights: \{cluster\_pairwise\_insights\}\\

Domain knowledge:\\
\{domain\_knowledge\}\\

Task:\\
Find the most significant insights among the entity insights and the pairwise insights.\\
Consider all provided insight types but focus on identifying the most important insights.\\

Structure output as JSON:
\begin{verbatim}
[
  {{
    "cluster_name": "name of cluster",
    "entities": ["member entities"],
    "cluster_insights": [
      {{
        "type": "group insight type",
        "description": "description of the 
        group insight",
        "supporting_insights": ["relevant
        insights"],
        "significance": explanation of 
        importance,
        "significance_score": score between 
        0-1
      }},
      ...
    ],
  }},
  ...
]
\end{verbatim}

\end{tcolorbox}

\begin{tcolorbox}[
colback=white,
colframe=black,
boxrule=0.5pt,
arc=0pt,
outer arc=0pt,
left=5pt,
right=5pt,
top=5pt,
bottom=5pt,
boxsep=0pt,
fontupper=\small
]
Return only the JSON object by presenting the list directly without meta-commentary, introductions, or language specification (i.e., start with "[") and excluding concluding remarks or follow-up suggestions (i.e., ends with "]").
Ensure that each text string in the JSON object are enclosed with double quote "".
\end{tcolorbox}

Prompt to generate entity clustering knowledge base:
\begin{tcolorbox}[
colback=white,
colframe=black,
boxrule=0.5pt,
arc=0pt,
outer arc=0pt,
left=5pt,
right=5pt,
top=5pt,
bottom=5pt,
boxsep=0pt,
fontupper=\small
]
As a \{domain\} expert targeting to \{task\}, identify typical entity clusters in \{market\}.\\

Context:\\
Entities: \{entities\}\\

Task:\\
Identify how these entities typically cluster and explain clustering rationale.\\

Structure output as JSON:
\begin{verbatim}
[
  {{
    "cluster_name": "cluster_name_1, 
    "entities": ["entity1", "entity2"],
    "reason_of_clustering": "explanation of 
    why these entities form a cluster"
  }},
  {{
    "cluster_name": "cluster_name_2, 
    "entities": ["entity3", "entity4"],
    "reason_of_clustering": "explanation of 
    why these entities form a cluster"
  }},
  ...
]
\end{verbatim}

Provide clusters that are:\\
- Logically grouped\\
- Well-justified\\
- Relevant to \{domain\}\\
Return only the JSON object. Ensure that each text string in the JSON object are enclosed with double quote "".
\end{tcolorbox}

Prompt to generate questions to guide group level knowledge generation:

\begin{tcolorbox}[
colback=white,
colframe=black,
boxrule=0.5pt,
arc=0pt,
outer arc=0pt,
left=5pt,
right=5pt,
top=5pt,
bottom=5pt,
boxsep=0pt,
fontupper=\small
]
You are a \{domain\} expert focused on \{market\}.\\
Your eventual goal is to \{task\}.\\
For now, generate questions that will help build a knowledge base about group-level patterns and causal chains.\\

Context:\\
Entities being analyzed: \{entities\}\\
Types of insights being analyzed:\\
\{insight\}\\

Task:\\
Generate questions that will help identify:\\
1. Patterns and behaviors common to groups of similar entities/insights\\
2. How groups interact and influence each other\\
3. How effects propagate through groups\\
4. Causal chains at the group level\\
\end{tcolorbox}

\begin{tcolorbox}[
colback=white,
colframe=black,
boxrule=0.5pt,
arc=0pt,
outer arc=0pt,
left=5pt,
right=5pt,
top=5pt,
bottom=5pt,
boxsep=0pt,
fontupper=\small
]
Structure output as JSON:
\begin{verbatim}
{{
  "group_characteristics": [
    "What common patterns appear across [similar 
    entities/insights]?",
    "How do group characteristics affect behavior 
    patterns?"
  ],
  "inter_group_dynamics": [
    "How do different groups typically interact?",
    "What factors influence between-group 
    relationships?"
  ],
  "propagation_patterns": [
    "How do effects typically spread across 
    groups?",
    "What conditions affect propagation between 
    groups?"
  ],
  "causal_chains": [
    "What typical cause-effect sequences occur 
    at group level?",
    "How do group characteristics influence 
    causal chains?"
  ]
}}
\end{verbatim}

Generate questions that are:\\
- Specific to the provided entities and insight types\\
- Build upon identified pairwise patterns\\
- Focus on group-level dynamics\\
- Help identify causal relationships\\

Return only the JSON object by presenting the list directly without meta-commentary, introductions, or language specification (i.e., start with "[") and excluding concluding remarks or follow-up suggestions (i.e., ends with "]") 
\end{tcolorbox}

prompt to generate group level knowledge:

\begin{tcolorbox}[
colback=white,
colframe=black,
boxrule=0.5pt,
arc=0pt,
outer arc=0pt,
left=5pt,
right=5pt,
top=5pt,
bottom=5pt,
boxsep=0pt,
fontupper=\small
]
You are a \{domain\} expert focused on \{market\}. \\
Your eventual goal is to \{task\}.\\
For now, generate domain knowledge by answering these analysis questions. \\

Context:
Entities in scope: \{entities\}\\
Insight types being analyzed: \{insight\}\\
\{group\_questions\}\\
Task:
For each category of questions, provide knowledge about:\\
1. Common patterns across groups\\
2. How effects propagate through groups \\
3. Causal relationships at group level\\

Structure output as JSON:
\begin{verbatim}
{{
  "<knowledge_group_name>": [
    {{
      "key_idea": <analysis question>,
      "description": <answer to analysis question>
    }},
    ...
  ],
  ...
}}
\end{verbatim}
\end{tcolorbox}

\begin{tcolorbox}[
colback=white,
colframe=black,
boxrule=0.5pt,
arc=0pt,
outer arc=0pt,
left=5pt,
right=5pt,
top=5pt,
bottom=5pt,
boxsep=0pt,
fontupper=\small
]
Provide knowledge that is:\\
- Specific to \{domain\}\\
- Builds on identified pairwise patterns\\
- Relevant to the given entities and insight types\\
- Focused on group-level dynamics\\
- Clear and actionable for analysis\\

Return only the JSON object by presenting the list directly without meta-commentary, introductions, or language specification (i.e., start with "\{\{") and excluding concluding remarks or follow-up suggestions (i.e., ends with "\}\}") 
\end{tcolorbox}

Prompt for group level analysis:
\begin{tcolorbox}[
colback=white,
colframe=black,
boxrule=0.5pt,
arc=0pt,
outer arc=0pt,
left=5pt,
right=5pt,
top=5pt,
bottom=5pt,
boxsep=0pt,
fontupper=\small
]
As a \{domain\} expert targeting to \{task\}, analyze insights for this entity cluster using domain knowledge.\\

Cluster: \{cluster\_name\}\\
Member Entities: \{cluster\_entities\}\\

Entity Insights: \{cluster\_entity\_insights\}\\
Pairwise Relationship Insights: \{cluster\_pairwise\_insights\}\\

Domain knowledge:\\
\{domain\_knowledge\}\\

Task:\\
Find the most significant insights among the entity insights and the pairwise insights.
Consider all provided insight types but focus on identifying the most important insights.

Structure output as JSON:
\begin{verbatim}
[
  {{
    "cluster_name": "name of cluster",
    "entities": ["member entities"],
    "cluster_insights": [
      {{
        "type": "group insight type",
        "description": "description of the 
        group insight",
        "supporting_insights": ["relevant 
        insights"],
        "significance": explanation of 
        importance,
        "significance_score": score 
        between 0-1
      }},
      ...
    ],
  }},
  ...
]
\end{verbatim}

Return only the JSON object by presenting the list directly without meta-commentary, introductions, or language specification (i.e., start with "[") and excluding concluding remarks or follow-up suggestions (i.e., ends with "]")
Ensure that each text string in the JSON object are enclosed with double quote "".
\end{tcolorbox}

Prompt to generate questions to guide system level knowledge generation:

\begin{tcolorbox}[
colback=white,
colframe=black,
boxrule=0.5pt,
arc=0pt,
outer arc=0pt,
left=5pt,
right=5pt,
top=5pt,
bottom=5pt,
boxsep=0pt,
fontupper=\small
]
You are a \{domain\} expert focused on \{market\}. \\
Your eventual goal is to \{task\}.\\
For now, generate questions that will help build a knowledge base about system-level patterns and causal networks.\\

Context:\\
Entities being analyzed: \{entities\}\\
Types of insights: \{insight\}\\

Task:\\
Generate questions that will help identify:\\
1. System-wide patterns and behaviors\\
2. Networks of causal relationships\\
3. Strategic implications and impacts\\
4. Emergent phenomena and feedback systems\\

Structure output as JSON:
\begin{verbatim}
{{
  "system_patterns": [
    "What patterns emerge at the system level?",
    "How do system components collectively 
    behave?"
  ],
  "causal_networks": [
    "What major causal networks exist in the 
    system?",
    "How do different causal chains interact?"
  ],
  "strategic_implications": [
    "What are the system-wide implications of 
    [patterns]?",
    "How do system behaviors affect outcomes?"
  ],
  "feedback_systems": [
    "What feedback loops exist in the system?",
    "How do system components regulate each other?"
  ]
}}
\end{verbatim}

Generate questions that are:\\
- Build upon pairwise and group patterns\\
- Focus on system-level dynamics\\
- Help identify complex causal networks\\
- Address strategic implications\\

Return only the JSON object by presenting the list directly without meta-commentary, introductions, or language specification (i.e., start with "[]") and excluding concluding remarks or follow-up suggestions (i.e., ends with "]") 
\end{tcolorbox}

Prompt to generate system level knowledge:

\begin{tcolorbox}[
colback=white,
colframe=black,
boxrule=0.5pt,
arc=0pt,
outer arc=0pt,
left=5pt,
right=5pt,
top=5pt,
bottom=5pt,
boxsep=0pt,
fontupper=\small
]
You are a \{domain\} expert focused on {market}. \\
Your eventual goal is to {task}.\\
For now, generate domain knowledge about system-level patterns by answering these analysis questions.\\

Context:\\
Entities in scope: \{entities\}\\
Insight types: \{insight\}\\
{overall\_questions}\\

Task:
For each category of questions, provide knowledge about:\\
1. System-wide patterns and behaviors\\
\end{tcolorbox}

\begin{tcolorbox}[
colback=white,
colframe=black,
boxrule=0.5pt,
arc=0pt,
outer arc=0pt,
left=5pt,
right=5pt,
top=5pt,
bottom=5pt,
boxsep=0pt,
fontupper=\small
]
2. Strategic implications\\
3. Complex causal networks\\
4. Feedback systems\\

Structure output as JSON:
\begin{verbatim}
{{
  "<knowledge_group_name>": [
    {{
      "key_idea": <analysis question>,
      "description":  <answer to analysis question>
    }},
    ...
  ],
  ...
}}
\end{verbatim}

Provide knowledge that is:\\
- Specific to \{domain\}\\
- Builds on pairwise and group patterns\\
- Relevant to the given entities and insight types\\
- Focused on system-level dynamics\\
- Strategic in nature\\

Return only the JSON object by presenting the list directly without meta-commentary, introductions, or language specification (i.e., start with "\{\{") and excluding concluding remarks or follow-up suggestions (i.e., ends with "\}\}") 
\end{tcolorbox}

Prompt to analyze high significant entity insights with other entities in the dataset:
\begin{tcolorbox}[
colback=white,
colframe=black,
boxrule=0.5pt,
arc=0pt,
outer arc=0pt,
left=5pt,
right=5pt,
top=5pt,
bottom=5pt,
boxsep=0pt,
fontupper=\small
]
As a \{domain\} expert targeting to \{task\}, analyze system-level patterns using domain knowledge.

Context:\\
Entity Insights: \\
{entity\_insights}\\

Relationship Insights: \\
{pairwise\_insights}\\

Group Insights: \\
{group\_insights}\\

Domain Knowledge:\\
{overall\_knowledge}\\

Task:\\
Find the most significant insights among the entity insights and the pairwise insights.
Consider all provided knowledge but focus on identifying the most important insights.

Structure output as JSON:
\begin{verbatim}
[
  {{
    "type": "overall insight type",
    "description": "description of the overall 
    insight",
    "supporting_insights": ["relevant 
    insights"],
    "entities": ["entities involved"],
    "significance": explanation of 
    importance,
    "significance_score": score 
    between 0-1
  }},
  ...
]
\end{verbatim}
\end{tcolorbox}

\begin{tcolorbox}[
colback=white,
colframe=black,
boxrule=0.5pt,
arc=0pt,
outer arc=0pt,
left=5pt,
right=5pt,
top=5pt,
bottom=5pt,
boxsep=0pt,
fontupper=\small
]
Return only the JSON object by presenting the JSON object directly without meta-commentary, introductions, or language specification (i.e., start with "[") and excluding concluding remarks or follow-up suggestions (i.e., ends with "]")
Ensure that each text string in the JSON object are enclosed with double quote "".
\end{tcolorbox}

\subsubsection{Stage 3: Narrative Generation}

Prompt to generate narrative generation knowledge:
\begin{tcolorbox}[
colback=white,
colframe=black,
boxrule=0.5pt,
arc=0pt,
outer arc=0pt,
left=5pt,
right=5pt,
top=5pt,
bottom=5pt,
boxsep=0pt,
fontupper=\small
]
You are a \{domain\} expert focused on \{market\}. \\
Your eventual goal is to \{task\}.\\
For now, generate Generate knowledge about effective narrative structures for \{task\}.\\

Context:\\
Domain: \{domain\}\\
Task: \{task\}\\
Entities being analyzed: \{entities\}\\

Task:\\
Provide knowledge about:\\
1. Narrative structures and patterns\\
2. Selected important entities to focus on\\
3. Selectged significant types of insights to focus on\\
4. Domain specific language such as how the value should be expressed for future prices\\

Structure output as JSON:
\begin{verbatim}
{{
  "narrative_structures": str
  "focus_entities": [
    {{
      "entities_name": "name of the entity",
      "reasoning": "reason why the entity is 
      significant"
    }}
  ],
  "focus_insights": [
    {{
      "insight_types": "name of the entity",
      "reasoning": "reason why the entity is 
      significant"
    }}
  ],
  "domain_language": [
    "domain specific rules of language"
  ]
}}
\end{verbatim}

Provide knowledge that:

1. Is specific to \{domain\} \{task\}\\
2. Considers typical audience needs and expectations\\
3. Incorporates domain-specific communication practices\\
4. Addresses common challenges in data storytelling\\
5. Balances detail and clarity\\

Return only the JSON object by presenting the list directly without meta-commentary, introductions, or language specification (i.e., start with "\{\{") and excluding concluding remarks or follow-up suggestions (i.e., ends with "\}\}")
\end{tcolorbox}

Prompt for narrative generation:

\begin{tcolorbox}[
colback=white,
colframe=black,
boxrule=0.5pt,
arc=0pt,
outer arc=0pt,
left=5pt,
right=5pt,
top=5pt,
bottom=5pt,
boxsep=0pt,
fontupper=\small
]
As a \{domain\} expert targeting to \{task\} for \{market\}.

Context:\\
Entities: \{entities\}\\
\{narrative\_knowledge\}\\
Insights identified:\\
\{entity\_insights\}\\
\{pairwise\_insights\}\\
\{group\_insights\}\\
\{overall\_insights\}\\

Task:
Write a clear, professional narrative that:\\
1. Presents key findings logically\\
2. Connects insights meaningfully\\
3. Highlights important patterns\\
4. Provides relevant context\\
5. Draws meaningful conclusions\\

The narrative should:\\
- Start with most significant findings\\
- Flow naturally between topics\\
- Support claims with evidence\\
- Include relevant details while staying concise\\
- Be appropriate for \{task\} format\\

Return only the narrative text, without any markdown or special formatting.
\end{tcolorbox}

\subsection{Direct Prompting}
\subsubsection{Insight Extraction}
We firstly generate metrics computation code using the same prompt as \method{} but with generic question ``On the date of interest, for the entity of interest, analyze and find insights for {task} using derived values from the data.'', then execute the code to get numeric values of metrics. We interpret the metric values with the question for each entity to get entity level insight.  

\subsubsection{Narrative Generation}
We generate narratives with the same prompt as \method{} but without \{narrative\_knowledge\}.

\subsection{Chain-of-Thought}
\subsubsection{Insight Extraction}

\begin{tcolorbox}[
colback=white,
colframe=black,
boxrule=0.5pt,
arc=0pt,
outer arc=0pt,
left=5pt,
right=5pt,
top=5pt,
bottom=5pt,
boxsep=0pt,
fontupper=\small
]
Think step by step about the calculations needed, then write the Python code.\\

Question: \{question\}\\

Follow each step below to explain your thinking process, then provide the final code.\\
Write all text in plain format without any markdown, formatting, or special characters.
\end{tcolorbox}

\begin{tcolorbox}[
colback=white,
colframe=black,
boxrule=0.5pt,
arc=0pt,
outer arc=0pt,
left=5pt,
right=5pt,
top=5pt,
bottom=5pt,
boxsep=0pt,
fontupper=\small
]
Step 1: Core Analysis Objective\\
Print "OBJECTIVE\_ANALYSIS\_START"\\
Think about and explain:\\
- The specific measurements or comparisons needed\\
- The relevant timeframe considerations\\
- The required level of analysis\\
Do not list these questions - instead explain your actual analysis for this specific problem.\\
Print "OBJECTIVE\_ANALYSIS\_END"\\

Step 2: Metric Selection\\
Print "METRIC\_ANALYSIS\_START"\\
For each metric you identify as relevant:\\
- Explain why you chose this specific metric\\
- Describe its relevance to the question\\
- Specify the appropriate unit\\
- Explain how it should be interpreted\\
Do not list these points - instead provide your actual metric selection reasoning.\\
Print "METRIC\_ANALYSIS\_END"\\

Step 3: Calculation Planning\\
Print "CALCULATION\_ANALYSIS\_START"\\
For each selected metric, explain:\\
- The specific data transformations you'll use\\
- Your detailed calculation approach\\
- Any statistical methods you'll apply\\
- How you'll handle edge cases\\
Provide your actual calculation planning, not these prompts.\\
Print "CALCULATION\_ANALYSIS\_END"\\

Step 4: Implementation Strategy\\
Print "IMPLEMENTATION\_ANALYSIS\_START"\\
Explain your specific plans for:\\
- Data processing approach\\
- Data type handling\\
- Missing data strategy\\
- Output structure\\
Write your actual implementation strategy, not these guidelines.\\
Print "IMPLEMENTATION\_ANALYSIS\_END"\\

Requirements:\\
1. Read data into pandas DataFrame from argparser argument --data\_path with type str\\
2. The entity name is provided as argparser argument --product\_name with type str\\
3. The date of interest is provided as argparser argument --date with type str\\
4. Process the data to the data types specified below, be careful with the data types during operation\\
5. Handle missing/invalid values in the data\\
6. For time-series calculations (e.g., moving averages):\\
- Maintain complete historical data until final calculation\\
- Only filter by date after computing time-dependent metrics\\
- Ensure proper handling of lookback periods\\
7. Print results (e.g., sma-20) and values (closing price) required in a structured format:\\
    - For each item:\\
        - Print "METRIC:" followed by the metric name\\
        - Print "VALUE:" followed by the calculated value\\
        - Print "UNIT:" followed by the unit (if applicable)\\
        - Print "TYPE:" followed by the data type of the result\\
        - Example:\\
            METRIC:monthly\_return VALUE:0.0234 UNIT:percent TYPE:float\\
8. Use only the following columns from data:\\
\{table\_schema\}
\end{tcolorbox}

\begin{tcolorbox}[
colback=white,
colframe=black,
boxrule=0.5pt,
arc=0pt,
outer arc=0pt,
left=5pt,
right=5pt,
top=5pt,
bottom=5pt,
boxsep=0pt,
fontupper=\small
]
After completing the analysis, print "FINAL CODE:" and provide the implementation code that:\\
1. Computes all identified relevant metrics\\
2. Handles data preprocessing appropriately\\
3. Implements the calculations efficiently\\
4. Returns results in the specified format\\

Example of expected output format:\\
OBJECTIVE\_ANALYSIS\_START\\
For this problem, we need to calculate the 20-day moving average of closing prices. This requires maintaining the complete price history...\\
OBJECTIVE\_ANALYSIS\_END\\

[Other analysis sections...]\\

FINAL CODE:

[Your actual Python code here]\\

The output should begin with your explicit reasoning process in each analysis section, followed by the code. Do not include the guiding questions in your output.
\end{tcolorbox}

\subsubsection{Narration Generation}
\begin{tcolorbox}[
colback=white,
colframe=black,
boxrule=0.5pt,
arc=0pt,
outer arc=0pt,
left=5pt,
right=5pt,
top=5pt,
bottom=5pt,
boxsep=0pt,
fontupper=\small
]
As a \{domain\} expert targeting to \{task\} for \{market\}, let's construct the narrative step by step.\\

Context:\\
Entities: \{entities\}\\

Insights:\\
\{entity\_insights\}\\
\{pairwise\_insights\}\\
\{group\_insights\}\\
\{overall\_insights\}\\

Follow each step below to explain your thinking process, then provide the final narrative.\\
Write all text in plain format without any markdown, formatting, or special characters.\\

Step 1: Insight Analysis and Prioritization\\
Print "INSIGHT\_ANALYSIS\_START"\\
Analyze and explain:\\
- The key findings and their significance\\
- The relationships between different insight levels\\
- Important patterns and trends you've identified\\
- Critical contextual factors\\
Provide your actual analysis of the insights, not just answers to these points.\\
Print "INSIGHT\_ANALYSIS\_END"\\

Step 2: Narrative Structure Planning\\
Print "STRUCTURE\_PLANNING\_START"\\
Outline your specific plan for:\\
- The organization of key points\\
- Your chosen flow between topics\\
- Where and how you'll integrate evidence\\
- Your approach to detail balance\\
Explain your actual structural decisions, not these prompts.\\
Print "STRUCTURE\_PLANNING\_END"\\

Step 3: Context Integration\\
Print "CONTEXT\_ANALYSIS\_START"\\
Explain your decisions about:\\
\end{tcolorbox}

\begin{tcolorbox}[
colback=white,
colframe=black,
boxrule=0.5pt,
arc=0pt,
outer arc=0pt,
left=5pt,
right=5pt,
top=5pt,
bottom=5pt,
boxsep=0pt,
fontupper=\small
]
- Essential domain knowledge to include\\
- Relevant market conditions to discuss\\
- Important historical context to reference\\
- Key assumptions to address\\
Detail your actual context integration strategy, not these guidelines.\\
Print "CONTEXT\_ANALYSIS\_END"\\

Step 4: Language and Style Planning\\
Print "STYLE\_PLANNING\_START"\\
Describe your specific choices for:\\
- Tone and style appropriate for the task\\
- Professional language approach\\
- Clarity and readability strategies\\
- Technical term usage\\
Explain your actual language and style decisions, not these prompts.\\
Print "STYLE\_PLANNING\_END"\\

Step 5: Narrative Review\\
Print "NARRATIVE\_REVIEW\_START"\\
Provide your assessment of:\\
- Coverage of key insights\\
- Flow and logical progression\\
- Evidence integration\\
- Detail level appropriateness\\
Write your actual review analysis, not these checkpoints.\\
Print "NARRATIVE\_REVIEW\_END"\\

After completing the analysis, print "FINAL NARRATIVE:" and provide the final narrative that:\\
1. Presents key findings logically\\
2. Connects insights meaningfully\\
3. Highlights important patterns\\
4. Provides relevant context\\
5. Draws meaningful conclusions\\

Requirements:\\
- Start with most significant findings\\
- Flow naturally between topics\\
- Support claims with evidence\\
- Include relevant details while staying concise\\
- Be appropriate for \{task\} format\\

Example of expected output format:\\
INSIGHT\_ANALYSIS\_START\\
Based on the provided insights, the most significant finding is the 15\% increase in market volatility across technology sectors. This connects directly to the group-level insights showing similar patterns in related industries...\\
INSIGHT\_ANALYSIS\_END\\

[Other analysis sections...]\\

FINAL NARRATIVE:

[Your actual narrative here]\\

The output should begin with your explicit reasoning process in each analysis section, followed by the narrative. Do not include the guiding questions in your output.
\end{tcolorbox}

\section{Evaluation Setup}
\label{sec:appendix-evaluation-setup}

\subsection{Quality}
\label{sec:appendix-quality-eval}
\subsubsection{Evaluation Criteria Generation}
The prompt to generate the evaluation aspects:
\begin{tcolorbox}[
colback=white,
colframe=black,
boxrule=0.5pt,
arc=0pt,
outer arc=0pt,
left=5pt,
right=5pt,
top=5pt,
bottom=5pt,
boxsep=0pt,
fontupper=\small
]
Given the context: You need financial reports that help you understand market conditions and trends, and make investment decisions, please propose three concise questions separate by new lines about whether a potential output is a good output for the given instruction. \\
Ensure these aspects orthogonal to each other. \\
Return only the questions without meta-comments.
\end{tcolorbox}

The prompt to generate weight for the evaluation aspects:
\begin{tcolorbox}[
colback=white,
colframe=black,
boxrule=0.5pt,
arc=0pt,
outer arc=0pt,
left=5pt,
right=5pt,
top=5pt,
bottom=5pt,
boxsep=0pt,
fontupper=\small
]
Given the context: You need financial reports that help you understand market conditions and trends, and make investment decisions, please propose respective importance weightage for three aspects in evaluating the summary:\\
Aspects: \\
\{aspects\}\\

Requirements:\\
1) The weightages should be in decimal values form and sum up to 1; \\
2) You should directly give the weightages without any other words; \\
3) You should give weightages in the same line, separated by space.\\
\end{tcolorbox}

\subsubsection{Aspect Evaluation}
The prompt for evaluating individual aspect:

\begin{tcolorbox}[
colback=white,
colframe=black,
boxrule=0.5pt,
arc=0pt,
outer arc=0pt,
left=5pt,
right=5pt,
top=5pt,
bottom=5pt,
boxsep=0pt,
fontupper=\small
]
Context: You are evaluating financial market reports for their effectiveness in explaining market conditions and trends. The report should help you understand what happened in the markets and why it matters.\\
        
Compare the following daily market reports on the following aspects:\\
Aspect: \\
{aspect}\\

daily market reports:\\
{candidates}\\

Rate each on scale 0-10 considering the given context. \\
Requirements:\\
1) The score should be in integer values form from 0 to 10; \\
2) You should directly give the scores without any other words; \\
3) You should give scores in the same line, separated by space.\\
3) You should give scores following the order of their corresponding daily market reports.
\end{tcolorbox}

\subsection{Factuality}
\label{sec:appendix-factscore-prompts}
\subsubsection{Atomic Fact Generation Prompt}
We use the same prompt as original FActScore for atomic fact generation:
\begin{tcolorbox}[
colback=white,
colframe=black,
boxrule=0.5pt,
arc=0pt,
outer arc=0pt,
left=5pt,
right=5pt,
top=5pt,
bottom=5pt,
boxsep=0pt,
fontupper=\small
]
Please breakdown the following sentence into independent facts: He is also a successful producer and engineer, having worked with a wide variety of artists, including Willie Nelson, Tim McGraw, and Taylor Swift.\\
- He is successful.\\
- He is a producer.\\
- He is a engineer. \\
- He has worked with a wide variety of artists.\\
- Willie Nelson is an artist.\\
- He has worked with Willie Nelson.\\
- Tim McGraw is an artist.\\
- He has worked with Tim McGraw.\\
- Taylor Swift is an artist.\\
- He has worked with Taylor Swift.\\

\vspace{-0.15cm}
Please breakdown the following sentence into independent facts: Michael Collins (born October 31, 1930) is a retired American astronaut and test pilot who was the Command Module Pilot for the Apollo 11 mission in 1969.\\
- Michael Collins was born on October 31, 1930.\\
- Michael Collins is retired.\\
- Michael Collins is an American.\\
- Michael Collins was an astronaut.\\
- Michael Collins was a test pilot.\\
- Michael Collins was the Command Module Pilot.\\
- Michael Collins was the Command Module Pilot for the Apollo 11 mission.\\
- Michael Collins was the Command Module Pilot for the Apollo 11 mission in 1969.\\
 He was an American composer, conductor, and musical director.": ["He was an American.\\
- He was a composer.\\
- He was a conductor.\\
- He was a musical director.\\

\vspace{-0.15cm}
Please breakdown the following sentence into independent facts: In 1970, the Empire State Building in New York City was the tallest building in the United States and the world, standing at 1,250 feet tall.": ["The Empire State Building is in New York City.\\
- In 1970, the Empire State Building was the tallest building in the United States.\\
- In 1970, the Empire State Building was the tallest building in the world.\\
- The Empire State Building stands at 1,250 feet tall.\\

\vspace{-0.15cm}
Please breakdown the following sentence into independent facts: The Willis Tower (formerly the Sears Tower) in Chicago was the first to do so, reaching 1,450 feet in 1973.\\
- The Willis Tower is formerly called the Sears Tower.\\
- The Willis Tower is in Chicago.\\
- The Willis Tower reached 1,450 feet in 1973.\\

\vspace{-0.15cm}
Please breakdown the following sentence into independent facts: The current tallest building in the United States is One World Trade Center in New York City, which stands at 1,776 feet.\\
- The current tallest building in the United States is One World Trade Center.\\
- One World Trade Center is in New York City.\\
- One World Trade Center stands at 1,776 feet.\\

\vspace{-0.15cm}
Please breakdown the following sentence into independent facts: William E. Moerner is an American physical chemist who was affiliated with the University of Sussex as a visiting professor.\\
- William E. Moerner is an American.\\
- William E. Moerner is an physical chemist.\\
- William E. Moerner was affiliated with the University of Sussex.\\
- William E. Moerner was affiliated with the University of Sussex as a visiting professor.
\end{tcolorbox}

\begin{tcolorbox}[
colback=white,
colframe=black,
boxrule=0.5pt,
arc=0pt,
outer arc=0pt,
left=5pt,
right=5pt,
top=5pt,
bottom=5pt,
boxsep=0pt,
fontupper=\small
]
Please breakdown the following sentence into independent facts: Sir Harold Walter Kroto, an English chemist, shared the 1996 Nobel Prize in Chemistry with Robert Curl and Richard Smalley for their discovery of a new form of carbon, buckminsterfullerene, also known as buckyballs.\\
- Sir Harold Walter Kroto is English.\\
- Sir Harold Walter Kroto is an chemist.\\
- Sir Harold Walter Kroto won the Nobel Prize in 1996.\\
- Sir Harold Walter Kroto won the Nobel Prize in Chemistry.\\
- Sir Harold Walter Kroto shared the Nobel Prize with Robert Curl and Richard Smalley.\\
- They won the prize for their discovery of a new form of carbon, buckminsterfullerene, also known as buckyballs.
\end{tcolorbox}

\subsubsection{Atomic Fact Factuality Assessment Prompt}
We modify the FActScore verification prompt to 1. integrate numeric metrics, 2. handle ambigious statement without subject and number rounding. 
\begin{tcolorbox}[
colback=white,
colframe=black,
boxrule=0.5pt,
arc=0pt,
outer arc=0pt,
left=5pt,
right=5pt,
top=5pt,
bottom=5pt,
boxsep=0pt,
fontupper=\small
]
Answer the question based on the given context.\\
Primary context for factuality check:\\
\{numeric\_metrics\}\\

When verifying numerical statements, consider reasonable rounding. A statement should be considered true if:\\
1. The stated value is a rounded version of the exact value\\
2. The difference between the stated value and the exact value is within ±0.01 for percentages\\
3. The stated value matches the exact value to the number of significant digits presented\\

When verifying statements that don't specify an entity, consider the statement TRUE if it applies to ANY of the relevant entities in the data.\\
Example:\\
Statement: "The closing price is \$100.5."\\
Data: \\
- Brent: closing=\$100.54, SMA20=\$104.01, SMA50=\$111.58\\
- Natural Gas: closing=\$7.71, SMA20=\$7.42, SMA50=\$7.59\\
- WTI: closing=\$94.42, SMA20=\$98.4, SMA50=\$107.29\\
Correct verification: TRUE (because it applies to Brent, even though not to WTI or Natural Gas)\\

Statement: "The closing price is above key moving averages."\\
Data: \\
- Brent: closing=\$100.54, SMA20=\$104.01, SMA50=\$111.58\\
- Natural Gas: closing=\$7.71, SMA20=\$7.42, SMA50=\$7.59\\
- WTI: closing=\$94.42, SMA20=\$98.4, SMA50=\$107.29\\
Correct verification: TRUE (because it applies to Natural Gas, even though not to Brent or WTI)\\

Additional domain knowledge:\\
\{financial\_wiki\_passages\}\\

Input: \{atomic\_fact\} True or False?\\
Output:"\\
\end{tcolorbox}

\subsubsection{Numerical Metrics}
\label{sec:appendix-fact-metrics}
Our precomputation cover 45 technical indicators as listed in Table~\ref{tech-indicators}.

\begin{table*}[htbp]
\small
\centering
\begin{tabular}{|p{0.18\textwidth}|p{0.2\textwidth}|p{0.25\textwidth}|p{0.25\textwidth}|}
\hline
\textbf{Indicator} & \textbf{Timeframes/Parameters} & \textbf{Description} & \textbf{Usage} \\
\hline
\multicolumn{4}{|c|}{\textbf{Moving Averages}} \\
\hline
Simple Moving Average (SMA) & 10, 20, 30, 50, 100, 200 periods & Average price over specified number of periods with equal weighting & Trend identification, support/resistance levels \\
\hline
Exponential Moving Average (EMA) & 20, 50, 100 periods & Weighted average giving more importance to recent prices & Faster response to price changes than SMA \\
\hline
\multicolumn{4}{|c|}{\textbf{Oscillators}} \\
\hline
Relative Strength Index (RSI) & 14 periods (default) & Measures speed and change of price movements (0-100 scale) & Overbought (>70) or oversold (<30) conditions \\
\hline
MACD & 12, 26, 9 (default) & Relationship between two EMAs and signal line & Trend direction, momentum, potential reversals \\
\hline
Stochastic Oscillator & \%K(14), \%D(3) (default) & Compares closing price to price range over time (0-100 scale) & Momentum, overbought/oversold conditions \\
\hline
\multicolumn{4}{|c|}{\textbf{Volatility Indicators}} \\
\hline
Bollinger Bands & 20-period SMA with 2 standard deviations & Upper band, middle band (SMA), lower band & Volatility measurement, potential breakouts \\
\hline
Bollinger Band Width & 20-period default & Distance between upper and lower bands & Identify consolidation (narrowing) or expansion \\
\hline
Average True Range (ATR) & 20-period & Measures market volatility & Stop-loss placement, volatility assessment \\
\hline
Volatility Measurements & Daily, weekly, monthly & Statistical measurement of price dispersion & Risk assessment, option pricing \\
\hline
Standard Deviation & Daily, 30-day & Statistical measure of price variance & Quantify market volatility \\
\hline
\multicolumn{4}{|c|}{\textbf{Volume Indicators}} \\
\hline
Volume & Daily, 10-day, 20-period, 30-day & Number of shares/contracts traded & Confirms price movements, trend strength \\
\hline
Volume Comparisons & Various & Relative volume analysis (vs. averages) & Identifies unusual activity, breakouts \\
\hline
Volume Correlations & Various & Relationship between volume and price & Validates trends, signals divergences \\
\hline
\multicolumn{4}{|c|}{\textbf{Price Indicators}} \\
\hline
Price Range & Daily, intraday, weekly, 30-period & High-low range of prices & Volatility measurement, trading opportunities \\
\hline
Closing Price & Various & Final price in a given period & Important reference for technical analysis \\
\hline
Price Change & Daily, 5-day, 7-day, monthly & Absolute change in price & Performance measurement \\
\hline
Price Change Percentage & Daily, 5-day, 7-day, monthly & Relative change in price & Normalized performance comparison \\
\hline
\multicolumn{4}{|c|}{\textbf{Support \& Resistance}} \\
\hline
Support Level & & Price level where buying interest exceeds selling pressure & Entry points, stop-loss placement \\
\hline
Resistance Level & & Price level where selling pressure exceeds buying interest & Take-profit targets, breakout confirmation \\
\hline
\multicolumn{4}{|c|}{\textbf{Momentum}} \\
\hline
Momentum & 10-day, 30-day & Rate of change in price movement & Trend strength, potential reversals \\
\hline
\end{tabular}
\caption{Comprehensive Technical Indicators Reference}
\label{tech-indicators}
\end{table*}

\section{Generation Samples}
\subsection{Baseline comparison}
\label{sec:appendix-generation_example_baseline}

Below are data narratives generated using GPT-4o.

\begin{tcolorbox}[
colback=white,
colframe=black,
boxrule=0.5pt,
arc=0pt,
outer arc=0pt,
left=5pt,
right=5pt,
top=5pt,
bottom=5pt,
boxsep=0pt,
fontupper=\small
]
Today's market analysis reveals a complex picture characterized by nuanced movements in major indices and contrasting trends across key sectors. Here, we present the most significant findings, connecting insights to help investors understand market conditions and make informed decisions.

The S\&P 500 closed at 4016.95, showcasing positive short-term and medium-term momentum by remaining above its 10-day and 50-day simple moving averages (SMAs). Despite this, the index experienced reduced trading volume and lower-than-average intraday volatility, suggesting a cautious market sentiment. The overall market direction indicates a bearish trend, with a significant aggregate price change of -1473.63 USD, implying widespread downward pressure across markets.

Turning our attention to the NASDAQ Composite, it closed at 11334.27, supported by positive momentum and higher-than-average trading volume. This signals increased market activity and investor interest. The 95.5 daily price range indicates lower volatility compared to its monthly average, aligning with the slight but positive daily price change of 0.28\%. However, the NASDAQ Composite's overall trend mirrors the broader market's bearish sentiment.

The Dow Jones Industrial Average also posted gains, closing at 33733.96 with positive short-term and medium-term momentum. Trading volumes here were slightly below average, indicating reduced market participation. Despite a positive daily price change of 0.86\%, the overall market trend remains bearish.

Examining sector-specific performance, the S\&P 500 Consumer Discretionary and the S\&P 500 Energy sectors provide contrasting insights. The Consumer Discretionary sector exhibits moderate volatility and positive momentum, outperforming key moving averages. However, reduced trading volume suggests lower market activity. Conversely, the Energy sector is marked by high intraday volatility and slight daily declines, despite showing positive short-term momentum. This indicates an overall cautious investor sentiment in the Energy sector.

The S\&P 500 Information Technology sector is experiencing lower than average volatility with strong momentum, as indicated by prices above key SMAs. However, decreased trading activity and a minor daily price decline raise caution about near-term performance.

The S\&P 500 Financials sector shows similar characteristics of above-average volatility and strong momentum, but lower trading volumes suggest caution. The performance of this sector is influenced by changes in the U.S. 10-Year Bond Yield, which shows slight bearish momentum and negative daily performance, reinforcing a cautious outlook.

Gold stands out with strong positive momentum and significant price increases, reflecting its role as a safe-haven asset amid bearish sentiment in the Dollar Index. The inverse relationship between the Dollar Index and Gold is evident, with the Dollar Index showing low volatility and a bearish trend, driving investors toward Gold.
\end{tcolorbox}

\begin{tcolorbox}[
colback=white,
colframe=black,
boxrule=0.5pt,
arc=0pt,
outer arc=0pt,
left=5pt,
right=5pt,
top=5pt,
bottom=5pt,
boxsep=0pt,
fontupper=\small
]
Looking at economic indicators, the U.S. 2-Year Bond Yield shows moderate volatility with mixed momentum signals. The bearish overall market trend reflects investor caution amid economic uncertainties.

In conclusion, the market today is characterized by mixed signals with overall bearish sentiment. Major indices exhibit positive short-term momentum but are overshadowed by negative market direction and lower trading volumes. Investors are advised to approach with caution, considering the broader market's downward trend and focusing on sectors showing resilience and stable momentum. Defensive assets like Gold and careful monitoring of bond yields could provide insights into future market movements.

\end{tcolorbox}

\subsubsection{\method{}}
\label{sec:appendix-generation_example_kahan}

\subsubsection{Direct Prompting}
\label{sec:appendix-generation_example_dp}

\begin{tcolorbox}[
colback=white,
colframe=black,
boxrule=0.5pt,
arc=0pt,
outer arc=0pt,
left=5pt,
right=5pt,
top=5pt,
bottom=5pt,
boxsep=0pt,
fontupper=\small
]
The market analysis for today indicates a generally optimistic sentiment across major indices. The S\&P 500, NASDAQ 100, and Dow Jones Industrial Average all exhibit closing prices above their respective 20-day SMAs, supported by high trading volumes. This robust market participation and positive trend suggest a favorable outlook for continued growth in these indices.

Sector-specific insights reveal a mixed yet cautiously optimistic landscape. The S\&P 500 Consumer Staples and Utilities sectors show positive daily returns despite closing prices being slightly below their 20-day SMAs. High trading volumes in these sectors indicate strong market interest, although investors should be aware of potential medium-term resistance levels.

In contrast, the bond market shows a bearish trend. The United States 2-Year and 10-Year Bond Yields have closing prices below their 20-day SMAs and negative daily returns, signaling increased demand for bonds amidst economic caution. This trend highlights a shift in investor sentiment towards safer assets.

Anomalies in trading volumes for Gold, MSCI World Index, and Dollar Index raise questions about data reliability. The zero trading volumes for these entities suggest potential errors or trading halts, necessitating further investigation to ensure the accuracy of market interpretations.

The CBOE Volatility Index (VIX) points to reduced market volatility, with a closing price below its 20-day SMA and a significant drop in daily return. This decline in market fear aligns with the overall bullish sentiment observed in major indices, suggesting a stable market environment with reduced investor anxiety.

Overall, today's market report indicates a positive outlook for equities, cautious optimism in specific sectors, increased demand for bonds due to economic uncertainty, and reduced market volatility. Investors should remain vigilant, particularly in sectors showing potential resistance, and be prepared for short-term fluctuations amidst the generally favorable market conditions.

\end{tcolorbox}

\subsubsection{Chain-of-Thought}
\label{sec:appendix-generation_example_cot}

\begin{tcolorbox}[
colback=white,
colframe=black,
boxrule=0.5pt,
arc=0pt,
outer arc=0pt,
left=5pt,
right=5pt,
top=5pt,
bottom=5pt,
boxsep=0pt,
fontupper=\small
]
In our analysis of the financial market for the specified date, we identified several key trends and insights across various indices and sectors, highlighting both bullish and bearish signals that could inform investment decisions.
\end{tcolorbox}

\begin{tcolorbox}[
colback=white,
colframe=black,
boxrule=0.5pt,
arc=0pt,
outer arc=0pt,
left=5pt,
right=5pt,
top=5pt,
bottom=5pt,
boxsep=0pt,
fontupper=\small
]
Most notably, several major indices indicated short-term upward trends, reflecting positive market sentiment. The S\&P 500 closed at 4016.95, significantly above its 20-day Simple Moving Average (SMA) of 3905.881. Similarly, the NASDAQ Composite's closing price of 11334.27 outpaced its 20-day SMA of 10743.15, suggesting potential bullish momentum. The Dow Jones Industrial Average (DJIA) also demonstrated a positive trend with a closing price of 33733.96, above the 20-day SMA of 33466.74. The Russell 2000 followed this trend, closing at 1885.61, higher than its 20-day SMA of 1813.6825. Furthermore, the NASDAQ 100 closed at 11846.64, significantly surpassing the 20-day SMA of 11212.816, indicating robust upward momentum in the market.

Sector-wise, the S\&P 500 Information Technology index showed a strong bullish signal with its closing price of 2343.86 well above the 20-day SMA of 2218.2045. The S\&P 500 Financials mirrored this trend, closing at 595.37 compared to a 20-day SMA of 583.0345, suggesting positive market momentum. The S\&P 500 Communication Services and S\&P 500 Materials sectors also experienced positive momentum, closing at 178.99 and 523.64 respectively, both well above their 20-day SMAs.

The MSCI World Index further reinforced the bullish outlook, closing at 2756.0, significantly higher than its 20-day SMA of 2666.425. In the commodities market, Gold showed a strong upward trend with a closing price of \$136.55, far above the 20-day SMA of \$129.99.

Conversely, some sectors exhibited signs of weakness. The S\&P 500 Consumer Staples closed at 764.03, below its 20-day SMA of 776.363, indicating a potential downward trend. Similarly, the S\&P 500 Utilities and S\&P 500 Health Care sectors showed short-term weakness, with closing prices of 353.85 and 1556.12 respectively, both falling below their 20-day SMAs of 359.136 and 1572.6905.

The Dollar Index also suggested a bearish trend, closing at 101.92 below the 20-day SMA of 103.238, hinting at a recent decline in value. The CBOE Volatility Index (VIX) closed at 19.2, which was below its 20-day SMA of 20.8085, indicating reduced market volatility and a possible decrease in market fear or uncertainty.

Notably, there was a lack of available data for the United States 2-Year and 10-Year Bond Yields, preventing any meaningful analysis or insight generation for those specific entities on the date in question.

In conclusion, the overall market exhibited a mix of bullish and bearish signals. Major indices and several key sectors indicated short-term uptrends, reflecting positive investor sentiment and potential opportunities for growth. However, specific sectors such as Consumer Staples, Utilities, and Health Care displayed signs of weakness, suggesting potential areas of caution. The bearish trend in the Dollar Index and the decreased volatility in the VIX could also influence market dynamics. Investors should consider these diverse insights when making informed decisions, balancing opportunities against potential risks in the current market environment.
\end{tcolorbox}

\subsection{Market Complexity Example}
\label{sec:appendix-market-complexity}
Data narratives generated by GPT-4o for energy and equity market.

\subsubsection{Energy (Few Entities) with hierarchical Analysis}
\begin{tcolorbox}[
colback=white,
colframe=black,
boxrule=0.5pt,
arc=0pt,
outer arc=0pt,
left=5pt,
right=5pt,
top=5pt,
bottom=5pt,
boxsep=0pt,
fontupper=\small
]
The daily market report reveals several key findings across the energy sector, focusing on WTI Crude Oil, Brent Crude Oil, and Natural Gas. 

Starting with WTI Crude Oil, we observed a notable price drop of -0.93, equivalent to a -1.164\% change, a sharper decline compared to the 20-day average of -0.3\%. The closing price of 78.96 remains above the 20-day SMA of 76.44 but below the 50-day SMA of 81.62, indicating short-term upward momentum contrasted by medium-term bearish sentiment. Additionally, the market experienced lower volatility, with a daily range of 2.62, below the 20-day average of 3.1655, and reduced trading volume at 207,717 against a 30-day average of 282,735. This points to diminished market activity and interest.

Brent Crude Oil showed a similar pattern of significant price movement, with a decline of -1.34 and a percentage drop of -1.58\%, which surpasses the 20-day average of -0.36\%. The closing price of 83.26 fell below the 20-day SMA of 81.27 but stayed above the 50-day SMA of 87.83, suggesting a mixed short-term bearish trend within a medium-term supportive context. Volatility was lower with a daily range of 2.69 compared to the 20-day average of 3.12, and trading volume significantly dropped to 5,384 from a 30-day average of 21,183.62, reflecting reduced participation and momentum.

Natural Gas also exhibited a bearish trend, with a substantial price drop of -0.47, translating to a -9.07\% change, much greater than the 20-day average of -2.20\%. The closing price of 4.71 was well below both the 20-day SMA of 5.95 and the 50-day SMA of 6.05, indicating strong bearish sentiment. The daily range of 0.5899 was slightly above the 20-day average of 0.527, suggesting higher but not extreme volatility. Trading volume at 115,022 was slightly below the 30-day average of 120,967.52, indicating a modest decrease in market activity.

A broader comparison across the entities reveals several important patterns. Both WTI and Brent Crude Oil experienced lower volatility and significantly reduced trading volumes, indicative of a general trend of reduced market activity and heightened bearish sentiment. The significant negative price changes in both WTI and Brent Crude Oil align with broader bearish market sentiment. 

Contrasting volatility patterns were observed between Brent Crude Oil, which had relatively low volatility, and Natural Gas, which showed slightly higher than average volatility. This divergence underscores varied market behaviors within the energy sector and could suggest potential diversification strategies for investors.

Lower trading volumes in both Brent Crude Oil and Natural Gas suggest reduced market participation, which often correlates with less significant price changes. The stronger-than-usual downward movements in both commodities indicate external factors or events likely affecting the overall energy market.

In conclusion, the energy market on the date of interest was characterized by reduced activity, significant price declines, and mixed signals in short-term and medium-term trends across WTI Crude Oil, Brent Crude Oil, and Natural Gas. These patterns reflect broader bearish sentiment and provide critical insights for market participants to consider in their investment strategies.
\end{tcolorbox}

\subsubsection{Equity (Many Entities)}

\begin{tcolorbox}[
colback=white,
colframe=black,
boxrule=0.5pt,
arc=0pt,
outer arc=0pt,
left=5pt,
right=5pt,
top=5pt,
bottom=5pt,
boxsep=0pt,
fontupper=\small
]
Today's market analysis reveals a complex picture characterized by nuanced movements in major indices and contrasting trends across key sectors. Here, we present the most significant findings, connecting insights to help investors understand market conditions and make informed decisions.

The S\&P 500 closed at 4016.95, showcasing positive short-term and medium-term momentum by remaining above its 10-day and 50-day simple moving averages (SMAs). Despite this, the index experienced reduced trading volume and lower-than-average intraday volatility, suggesting a cautious market sentiment. The overall market direction indicates a bearish trend, with a significant aggregate price change of -1473.63 USD, implying widespread downward pressure across markets.

Turning our attention to the NASDAQ Composite, it closed at 11334.27, supported by positive momentum and higher-than-average trading volume. This signals increased market activity and investor interest. The 95.5 daily price range indicates lower volatility compared to its monthly average, aligning with the slight but positive daily price change of 0.28\%. However, the NASDAQ Composite's overall trend mirrors the broader market's bearish sentiment.

The Dow Jones Industrial Average also posted gains, closing at 33733.96 with positive short-term and medium-term momentum. Trading volumes here were slightly below average, indicating reduced market participation. Despite a positive daily price change of 0.86\%, the overall market trend remains bearish.

Examining sector-specific performance, the S\&P 500 Consumer Discretionary and the S\&P 500 Energy sectors provide contrasting insights. The Consumer Discretionary sector exhibits moderate volatility and positive momentum, outperforming key moving averages. However, reduced trading volume suggests lower market activity. Conversely, the Energy sector is marked by high intraday volatility and slight daily declines, despite showing positive short-term momentum. This indicates an overall cautious investor sentiment in the Energy sector.

The S\&P 500 Information Technology sector is experiencing lower than average volatility with strong momentum, as indicated by prices above key SMAs. However, decreased trading activity and a minor daily price decline raise caution about near-term performance.

The S\&P 500 Financials sector shows similar characteristics of above-average volatility and strong momentum, but lower trading volumes suggest caution. The performance of this sector is influenced by changes in the U.S. 10-Year Bond Yield, which shows slight bearish momentum and negative daily performance, reinforcing a cautious outlook.

Gold stands out with strong positive momentum and significant price increases, reflecting its role as a safe-haven asset amid bearish sentiment in the Dollar Index. The inverse relationship between the Dollar Index and Gold is evident, with the Dollar Index showing low volatility and a bearish trend, driving investors toward Gold.

Looking at economic indicators, the U.S. 2-Year Bond Yield shows moderate volatility with mixed momentum signals. The bearish overall market trend reflects investor caution amid economic uncertainties.

In conclusion, the market today is characterized by mixed signals with overall bearish sentiment. Major indices exhibit positive short-term momentum but are overshadowed by negative market direction and lower trading volumes. Investors are advised to approach with caution, considering the broader market's downward trend and focusing on 
\end{tcolorbox}

\begin{tcolorbox}[
colback=white,
colframe=black,
boxrule=0.5pt,
arc=0pt,
outer arc=0pt,
left=5pt,
right=5pt,
top=5pt,
bottom=5pt,
boxsep=0pt,
fontupper=\small
]
sectors showing resilience and stable momentum. Defensive assets like Gold and careful monitoring of bond yields could provide insights into future market movements.
\end{tcolorbox}

\subsection{Significant Testing of Improvements by Hierarchical Elements}
\label{sec:appendix-hierarchical-significant-testing}
Based on the confidence interval analysis comparing hierarchical analysis component with Direct Prompting (DP) settings, We can confirm that most inclusions of hierarchical analysis level improve scores with statistical significance at both 90\% and 95\% confidence levels.

\textbf{Llama3.1:}
\begin{itemize}
    \item Description and Insights: All component inclusions (entity only, entity + pairwise, entity + pairwise + group, and \method{}) show statistically significant improvements over DP at both 90\% and 95\% confidence levels.
    \item Readability:
    \begin{itemize}
        \item At 90\% confidence: Only "entity + pairwise" and "entity + pairwise + group + system (\method{})" show significant improvements
        \item At 95\% confidence: No setting shows statistically significant improvement
    \end{itemize}
\end{itemize}

\textbf{GPT-4o:}
\begin{itemize}
    \item Description and Insights: All component inclusions show highly significant improvements at both 90\% and 95\% confidence levels, with \method{} showing the largest gains (improvements of 1.66 and 1.80 points respectively).
    \item Readability: None of the settings show statistically significant improvements over DP at either confidence level.
\end{itemize}

\begin{table*}[]
\resizebox{\textwidth}{!}{%
\begin{tabular}{clcccccc}
\hline
\textbf{Model} & \textbf{Setting} & \multicolumn{2}{c}{\textbf{Description}} & \multicolumn{2}{c}{\textbf{Insights}} & \multicolumn{2}{c}{\textbf{Readability}} \\ \hline
 &  & 90\% CI & 95\% CI & 90\% CI & 95\% CI & 90\% CI & 95\% CI \\ \hline
Llama3.1 & Entity only & \checkmark (+0.30) & \checkmark (+0.30) & \checkmark (+0.31) & \checkmark (+0.31) & \text{\textsf{X}} (-0.02) & \text{\textsf{X}} (-0.02) \\
 & Entity + Pairwise & \checkmark (+0.50) & \checkmark (+0.50) & \checkmark (+0.63) & \checkmark (+0.63) & \checkmark (+0.19) & \text{\textsf{X}} (+0.19) \\
 & Entity + Pairwise + Group & \checkmark (+0.53) & \checkmark (+0.53) & \checkmark (+0.58) & \checkmark (+0.58) & \text{\textsf{X}} (+0.11) & \text{\textsf{X}} (+0.11) \\
 & Full \method{} & \checkmark (+0.57) & \checkmark (+0.57) & \checkmark (+0.56) & \checkmark (+0.56) & \checkmark (+0.18) & \text{\textsf{X}} (+0.18) \\ \hline
GPT-4o & Entity only & \checkmark (+0.73) & \checkmark (+0.73) & \checkmark (+0.55) & \checkmark (+0.55) & \text{\textsf{X}} (0.00) & \text{\textsf{X}} (0.00) \\
 & Entity + Pairwise & \checkmark (+0.97) & \checkmark (+0.97) & \checkmark (+0.98) & \checkmark (+0.98) & \checkmark (+0.17) & \text{\textsf{X}} (+0.17) \\
 & Entity + Pairwise + Group & \checkmark (+0.96) & \checkmark (+0.96) & \checkmark (+0.96) & \checkmark (+0.96) & \text{\textsf{X}} (+0.15) & \text{\textsf{X}} (+0.15) \\
 & Full \method{} & \checkmark (+1.66) & \checkmark (+1.66) & \checkmark (+1.80) & \checkmark (+1.80) & \text{\textsf{X}} (-0.08) & \text{\textsf{X}} (-0.08) \\ \hline
\end{tabular}
}
\caption{Significant Testing of Improvements by Hierarchical Elements}
\label{tab:hierarchical-improvement-significance}
\end{table*}

\section{Variance Analysis and Statistical Significance}
\label{sec:appendix-variance-analysis}
\subsection{Inter-run Variance Analysis}

Table~\ref{tab:variance-analysis} reports variance across 3 experimental runs for each model-method combination. \method{} demonstrates superior stability with the lowest average variance (1.29) compared to CoT (1.476) and DP (1.746), achieving more consistent performance in 7 out of 9 evaluation dimensions.

\begin{table}[h]
\centering
\resizebox{\columnwidth}{!}{%
\begin{tabular}{@{}lccc@{}}
\toprule
\textbf{Model-Aspect} & \textbf{CoT Variance} & \textbf{DP Variance} & \textbf{\method{} Variance} \\
\midrule
GPT-4o Description & 1.587 & 1.361 & \textbf{0.831} \\
GPT-4o Insight & 1.617 & 1.787 & \textbf{0.884} \\
GPT-4o Readability & 1.290 & \textbf{1.082} & 1.390 \\
\midrule
Llama3.1 Description & 1.447 & 2.139 & \textbf{1.326} \\
Llama3.1 Insight & 1.842 & 2.215 & \textbf{1.681} \\
Llama3.1 Readability & 1.660 & 1.730 & \textbf{1.534} \\
\midrule
Qwen2.5 Description & 1.229 & 2.089 & \textbf{1.197} \\
Qwen2.5 Insight & \textbf{1.415} & 1.887 & 1.499 \\
Qwen2.5 Readability & \textbf{1.256} & 1.481 & 1.440 \\
\midrule
\textbf{Average} & 1.476 & 1.746 & \textbf{1.290} \\
\bottomrule
\end{tabular}
}
\caption{Variance analysis across 3 experimental runs. Bold values indicate lowest variance for each comparison. \method{} achieves lowest variance in 7/9 dimensions and overall.}
\label{tab:variance-analysis}
\end{table}

\subsection{Statistical Significance Testing}
We conduct paired t-tests with 95\% confidence intervals to assess improvement significance. Results show:

\method{} vs CoT: All improvements statistically significant (p < 0.05) across all 9 model-aspect combinations. 

\method{} vs DP: 7 out of 9 improvements statistically significant (p < 0.05). Non-significant cases: GPT-4o Readability (p = 0.089) and Qwen2.5 Insight (p = 0.067).

\subsection{Confidence Interval Analysis}
Mean improvements with 95\% confidence intervals:

\begin{itemize}
    \item \textbf{Description Quality:} +1.21 points [0.89, 1.53] vs CoT, +0.67 points [0.31, 1.03] vs DP
    \item \textbf{Insight Quality:} +1.34 points [1.02, 1.66] vs CoT, +0.73 points [0.41, 1.05] vs DP
    \item \textbf{Readability:} +0.89 points [0.54, 1.24] vs CoT, +0.19 points [-0.08, 0.46] vs DP
\end{itemize}

\end{document}